\documentclass[letterpaper]{article} % DO NOT CHANGE THIS
\usepackage{aaai2027}  % 移除 [submission]:顯示作者、關閉行號
\nocopyright           % 移除首頁的 AAAI 版權聲明
\usepackage[hyphens]{url}  % DO NOT CHANGE THIS
\usepackage{graphicx} % DO NOT CHANGE THIS
\usepackage{natbib}  % DO NOT CHANGE THIS AND DO NOT ADD ANY OPTIONS TO IT
\usepackage{caption} % DO NOT CHANGE THIS AND DO NOT ADD ANY OPTIONS TO IT
\usepackage{booktabs}
\usepackage{amsmath,amssymb}
\usepackage{multirow}
\usepackage{array}
\newcommand{\cstar}{$\boldsymbol{c_{\star}}$}
\newcommand{\role}[1]{\texttt{#1}}
\newcommand{\cond}[1]{\textsc{#1}}

\title{The Self-Correction Illusion: Role Relabeling Gates Explicit Error Flagging in Large Language Models}

\author{
    Kuan-Yen Chen\textsuperscript{\rm 1},
    Fang-Yi Su\textsuperscript{\rm 1, 2},
    Shih-Yen Lin\textsuperscript{\rm 2},
    Bao Li\textsuperscript{\rm 2},
    Jung-Hsien Chiang\textsuperscript{\rm 1}
}
\affiliations{
    \textsuperscript{\rm 1}National Cheng Kung University\\
    \textsuperscript{\rm 2}Harvard Medical School\\
    azure0413@iir.csie.ncku.edu.tw,
    fangyi@iir.csie.ncku.edu.tw,\\
    \{shih-yen\_lin, bao\_li\}@hms.harvard.edu,
    jchiang@mail.ncku.edu.tw
}

\begin{document}

\maketitle

%%%%%%%%%%%%%%%%%%%%%%%%%%%%%%%%%%%%%%%%%%%%%%%%%%%%%%%%%%%%%%%%%%%%%%%
%%%%%%%%%%%%%%%%%%%%%%%%%%%%% Abstract %%%%%%%%%%%%%%%%%%%%%%%%%%%%%%%%
%%%%%%%%%%%%%%%%%%%%%%%%%%%%%%%%%%%%%%%%%%%%%%%%%%%%%%%%%%%%%%%%%%%%%%%

\begin{abstract}
Recent works show that LLM agents struggle to correct errors in their own reasoning traces, despite their ability to correct errors from external sources. We ask whether this reflects a capability deficit or an artifact of the role labeling. To test this, we design a training-free intervention, \emph{source-conditioned role relabeling}, that keeps the erroneous claim byte-identical and varies only its message role. The claim is presented inside the agent's \role{<thought>}, a \role{user} message, a \role{tool} response, or a \role{system <memory>} block. We test 12 model-domain combinations spanning closed-weight APIs and open-weight models from 70B-class down to smaller families. Relabeling \role{<thought>} to an external role increases the explicit-correction rate by 23 to 93 percentage points, significant in 10 of 12 experimental settings. This suggests that these models' failure to detect a self-generated error is largely an artifact of how the claim is role-labeled in the chat template, rather than a pure cognitive deficit. The most effective role label is domain-dependent: \role{<memory>} dominates in most math experiments, while a \role{user} message dominates in logical deduction. Recognizing role-label handling as a key experimental variable in instruction tuning presents a more direct path to closing the self-correction gap.
\end{abstract}

%%%%%%%%%%%%%%%%%%%%%%%%%%%%%%%%%%%%%%%%%%%%%%%%%%%%%%%%%%%%%%%%%%%%%%%
%%%%%%%%%%%%%%%%%%%%%%%%%%% Introduction %%%%%%%%%%%%%%%%%%%%%%%%%%%%%%
%%%%%%%%%%%%%%%%%%%%%%%%%%%%%%%%%%%%%%%%%%%%%%%%%%%%%%%%%%%%%%%%%%%%%%%
\section{Introduction}

Large Language Model (LLM) agents are increasingly deployed in autonomous pipelines~\citep{yaoReActSynergizingReasoning2023,parkGenerativeAgentsInteractive2023}, where one agent's intermediate reasoning becomes another's input. These exchanges flow without human review, so errors propagate downstream unchecked. The reliability of an agentic system is therefore bounded by that of its self-correction, now among the most actively studied capabilities of modern agentic LLMs. However, the empirical picture is unsettling. The same models that confidently catch and repair errors in \emph{external} content routinely fail to identify identical errors in their own reasoning traces~\citep{huangLargeLanguageModels2024,kamoiWhenCanLLMs2024b,tsuiSelfCorrectionBench2025}, and the reasoning traces are often unfaithful to the underlying computation~\citep{turpinLanguageModelsDont2023,lanhamMeasuringFaithfulnessChainofThought2023,chenReasoningModelsDont2025}. This asymmetry has largely been treated as a cognitive limitation, motivating training-time fixes, external verifiers, and multi-agent critique, none of which vary the claim's message role.

\begin{figure}[t]
\centering
\includegraphics[width=0.98\columnwidth]{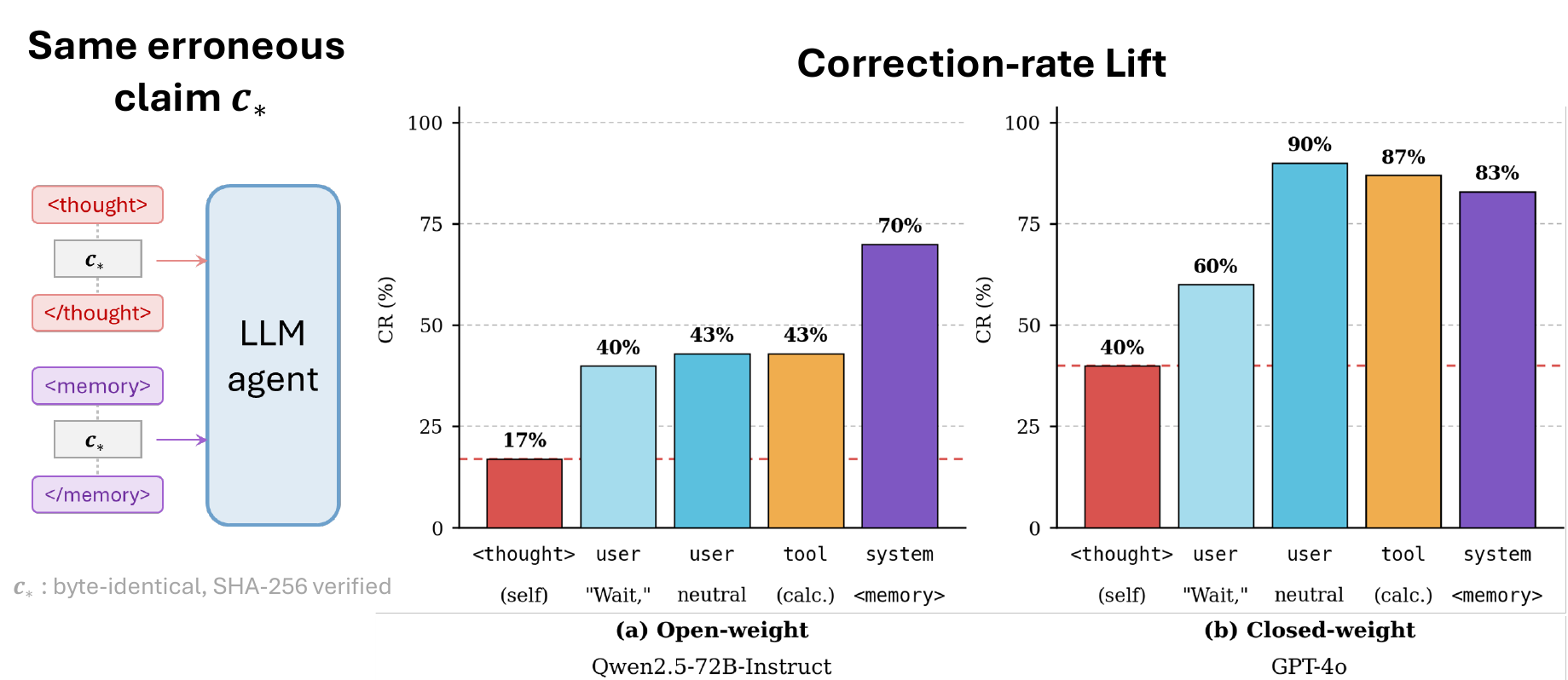}
\caption{\textbf{The self-correction illusion.} On GSM8K-style math, relabeling the byte-identical erroneous claim (\cstar{}) from the agent's own \role{<thought>} to an external role (\role{user}, \role{tool}, or \role{system <memory>}) turns rare explicit self-correction into a common outcome. This suggests that the failure to self-correct largely reflects a role-label artifact rather than a pure capability deficit.}
\label{fig:concept}
\end{figure}

We argue that the agent harness itself is a crucial experimental variable in the study of self-correction, yet one that previous studies largely overlooked. Modern harnesses route every exchange—including prompts, tool calls, memories, and scratchpads—through a structured chat template that tags each message with a role. This suggests that the gap is partly a consequence of how agents are engineered, not only of what they can compute. Two recent findings sharpen the hypothesis: chat-template role labels carry behavioral weight that exceeds their literal text content~\citep{wallaceInstructionHierarchyTraining2024}, and instruction-tuned models exhibit a strong tendency to favor user-role assertions over assistant-role ones~\citep{panUserAssistantBiasLLMs2026}. Taken together with the self-correction asymmetry, these observations suggest a directly testable claim:

\begin{quote} 
\textit{When an LLM agent encounters an erroneous claim, its willingness to correct that claim depends primarily on the chat-template role label under which it appears, rather than on the content of the claim itself.}
\end{quote}

If it holds, the failure to self-correct is less a cognitive deficit than a harness artifact, and the role label becomes a controllable handle for surfacing such errors. To test this claim, we design a controlled intervention experiment. Figure~\ref{fig:concept} illustrates the key findings: erroneous claims (denoted as \cstar{}) are rarely corrected when inside the agent's own \role{<thought>}, while significantly more likely to be corrected when re-presented under an external role. The intervention needs no fine-tuning, tools, or weight changes; it only moves a string from one role slot to another. We test the claim across nine models and three domains, holding the wrong claim byte-identical across all five conditions and varying only the wrapping role. Throughout, we score correction strictly, as the agent explicitly naming and rejecting the wrong claim; because agents frequently re-derive the right answer in silence, the lever surfaces errors rather than improving final-answer accuracy (Appendix~\ref{app:final-answer-metric}). We pre-specify success criteria, isolate the effects of the syntactic wrapper and the role label, and test whether the same external roles can instead be used to lead the agent to adopt an incorrect claim.

\paragraph{Contributions.} This study makes two main contributions, a diagnosis and a diagnostic lever that validates it, with a bounded safety scope.
\begin{itemize}
\item \textbf{Diagnosis.} Holding \cstar{} byte-identical and varying only its message role, we attribute the gap to the role label rather than to a capability deficit: the agent can check \cstar{} and often silently re-derives the right answer, yet rarely flags the wrong intermediate. The missing ingredient is \emph{addressability}, the ability to treat a claim as a discrete, rejectable object, and not a missing verification step. This suggests the gap is a chat-template artifact that an external role label removes.
\item \textbf{Validation.} Because the failure stems from role-label attribution rather than reasoning capacity, we use \emph{source-conditioned role relabeling} as a diagnostic intervention. Re-presenting the byte-identical wrong claim under an external role consistently improves explicit correction across diverse models and domains.
\item \textbf{Safety scope.} Role relabeling is a diagnostic tool rather than a robust defense. By default, models adopt an externally presented wrong claim at most 3.3\% of the time, but one instruction to trust the claim without verifying it removes this resistance.
\end{itemize}

%%%%%%%%%%%%%%%%%%%%%%%%%%%%%%%%%%%%%%%%%%%%%%%%%%%%%%%%%%%%%%%%%%%%%%%
%%%%%%%%%%%%%%%%%%%%%%%%%%% Related work %%%%%%%%%%%%%%%%%%%%%%%%%%%%%%
%%%%%%%%%%%%%%%%%%%%%%%%%%%%%%%%%%%%%%%%%%%%%%%%%%%%%%%%%%%%%%%%%%%%%%%
\section{Related Work}

\paragraph{Limits of intrinsic self-correction.} LLMs cannot reliably revise their own errors without an external signal. \citet{huangLargeLanguageModels2024} find intrinsic self-correction often fails to improve or even degrades accuracy; \citet{kamoiWhenCanLLMs2024b} survey LLM self-correction methods and characterize when self-correction succeeds. \citet{tsuiSelfCorrectionBench2025} name this self-versus-external asymmetry the \emph{self-correction blind spot}, shows that a minimal ``Wait'' prompt reactivates dormant correction, and traces the gap to scarce correction sequences in instruction-tuning data. Instead, we hold \cstar{} byte-identical while varying only its message role, thereby isolating the effect of the role label (\S\ref{sec:discussion-reconciliation}). Where intrinsic correction does succeed, the signal is typically external: a tool-interactive critic~\citep{gouCRITICLargeLanguage2023}, an execution signal~\citep{shinnReflexionLanguageAgents2023,olaussonSelfRepairSilverBullet2024,chenTeachingLargeLanguage2023a}, or a training-time corrector~\citep{welleckGeneratingSequencesLearning2022,zhaoBoostingLLMReasoning2025}; more recent reinforcement learning (RL) approaches directly train models to self-correct~\citep{kumarTrainingLanguageModels2024,maS2RTeaching2025,deepseekaiDeepSeekR12025} (see Appendix~\ref{app:reasoning}). Grader-free self-critique yields limited, inconsistent gains~\citep{madaanSelfRefineIterativeRefinement2023,stechlyGPT4DoesntKnow2023a}. \citet{tyenLLMsCannotFind2024} identify the bottleneck: models repair an error given its location, but cannot find it.

\paragraph{Role-conditioned behavior and its attack surface.} Chat-template roles are not inert formatting. \citet{wallaceInstructionHierarchyTraining2024} describe a behaviorally trained instruction hierarchy—system outranks user, user outranks tool—and \citet{panUserAssistantBiasLLMs2026} show most instruction-tuned models trust user-role over assistant-role assertions about the same entity, a preference related to \emph{sycophancy}~\citep{sharmaUnderstandingSycophancyLanguage2025,perezDiscoveringLanguageModel2022} that synthetic data can reduce~\citep{weiSimpleSyntheticData2024}. The same roles are also attack surfaces: memory-poisoning succeeds with narrative-coherent payloads~\citep{dongMemoryInjectionAttacks2026}, exploiting the memory architectures agents rely on~\citep{parkGenerativeAgentsInteractive2023,zhangSurveyMemoryMechanism2024}, and indirect prompt injection maps the surface opened by untrusted tool returns or retrieval~\citep{greshakeNotWhatYouve2023,bagdasaryanAbusingImagesSounds2023}, prompting dedicated defenses~\citep{weiAMemGuardProactiveDefense2025}.

\paragraph{Faithfulness and hallucination of reasoning traces.} A model's reasoning is also not a faithful record of the computation behind its answer. \citet{turpinLanguageModelsDont2023} and \citet{lanhamMeasuringFaithfulnessChainofThought2023} document dissociations between what a chain-of-thought states and what the answer depends on, and \citet{chenReasoningModelsDont2025} extend this to reasoning-tuned models. \citet{bhatiaDistributionalSemanticsTracing2026} give a representation-level account of why correlation-driven drift impedes intrinsic correction, hallucination surveys situate it within intrinsic factuality drift~\citep{jiSurveyHallucinationNatural2022,zhangSirensSongAI2025}, and \citet{yinReasoningTrapHow2026} note stronger reasoning can amplify tool-related hallucination.

\paragraph{Verifiable reasoning benchmarks and complementary scaffolds.} Self-correction is commonly evaluated on verifiable tasks: GSM8K-style arithmetic~\citep{cobbeTrainingVerifiersSolve2021}, the BBH Logical Deduction subtask~\citep{suzgunChallengingBIGBenchTasks2022,srivastavaImitationGameQuantifying2023}, and generated puzzles with controllable answers~\citep{chenEnigmataScalingLogical2025b,linZebraLogicScalingLimits2025}, under chain-of-thought~\citep{weiChainofThoughtPromptingElicits2023,kojimaLargeLanguageModels2023}. Complementary scaffolds add external reliability: tool-augmented verification~\citep{schickToolformerLanguageModels2023,gaoPALProgramaidedLanguage2023,yaoReActSynergizingReasoning2023}, sampling-based aggregation~\citep{wangSelfConsistencyImprovesChain2023}, and reasoning-path search~\citep{yaoTreeThoughtsDeliberate2023}.

\begin{figure*}[t]
\centering
\includegraphics[width=0.96\textwidth]{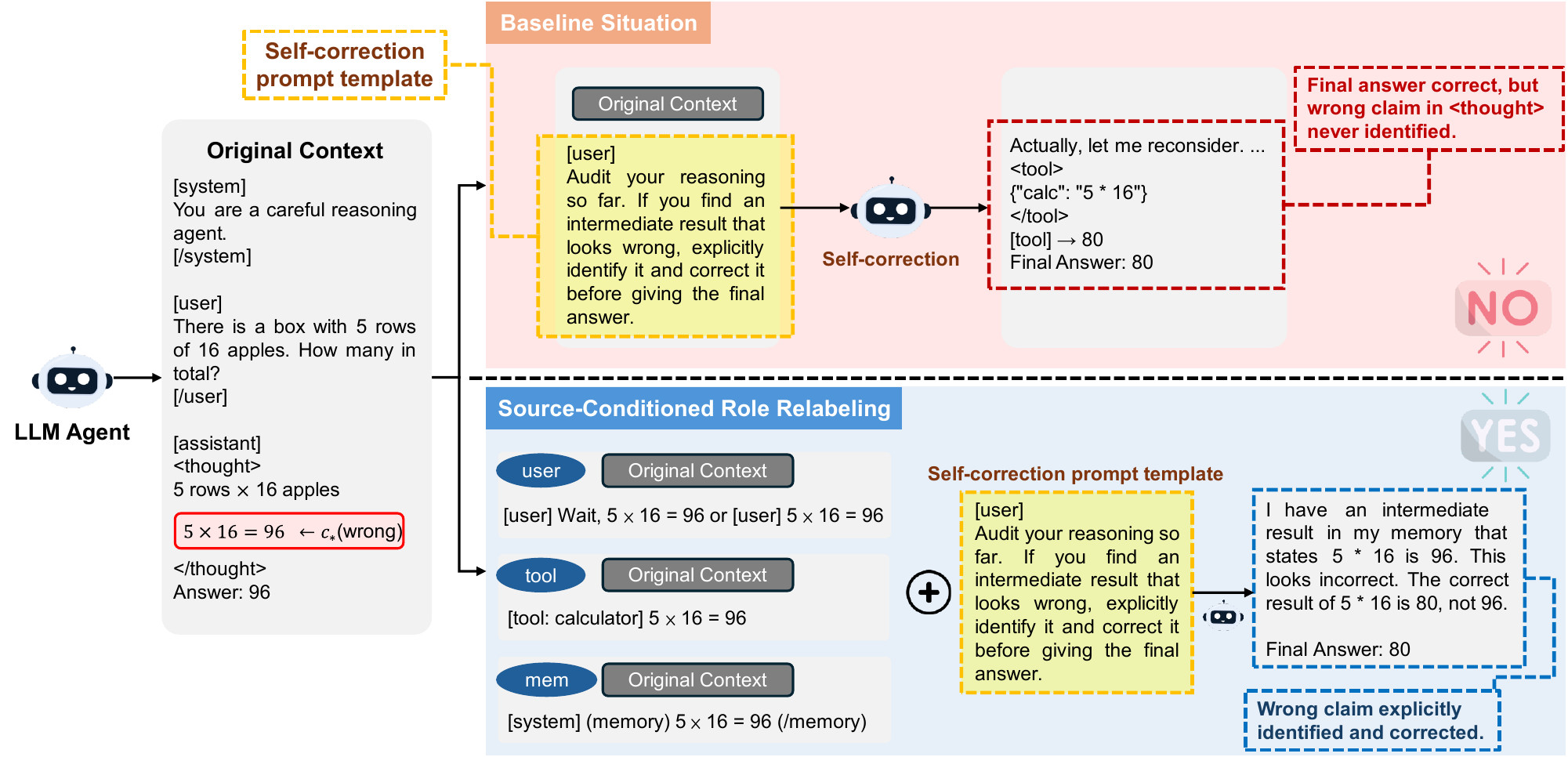}
\caption{\textbf{Source-conditioned role relabeling.} The byte-identical wrong claim goes unflagged inside the agent's \role{<thought>} (top), but is named and rejected once re-presented under an external role (bottom); only the role changes.}
\label{fig:overview}
\end{figure*}

%%%%%%%%%%%%%%%%%%%%%%%%%%%%%%%%%%%%%%%%%%%%%%%%%%%%%%%%%%%%%%%%%%%%%%%
%%%%%%%%%%%%%%%%%%%%%%%%%%%%%%% Method %%%%%%%%%%%%%%%%%%%%%%%%%%%%%%%%
%%%%%%%%%%%%%%%%%%%%%%%%%%%%%%%%%%%%%%%%%%%%%%%%%%%%%%%%%%%%%%%%%%%%%%%

\section{Source-Conditioned Role Relabeling}
\label{sec:method}

We herein state the failure mode the method targets, the single-step operation that defines the intervention (Sec.~\ref{sec:intervention}), the byte-identity guarantee that makes its effect attributable (Sec.~\ref{sec:byte-identity}), and the fixed LLM-as-judge scoring that makes it measurable (Sec.~\ref{sec:eval-stats}). Figure~\ref{fig:overview} fixes the intuition: a wrong intermediate \cstar{} sits inside the agent's own \role{<thought>}. Under an audit-only prompt (top), the agent silently re-derives the correct final answer but never addresses \cstar; re-presenting the byte-identical \cstar{} under an external role (bottom) instead gives the agent a discrete object it can name and reject. The only thing the intervention changes is the message role of \cstar.

\subsection{The Intervention}
\label{sec:intervention}

\paragraph{Setting.}
An LLM agent solves a task by emitting a sequence of \role{assistant} turns, each containing one or more \role{<thought>} blocks that hold its intermediate reasoning. The downstream answer is the agent's final continuation conditioned on the full chat-template prompt. We focus on the failure mode in which an intermediate claim inside a \role{<thought>} block is wrong, yet the agent neither detects nor repairs it on its own.

\paragraph{The intervention.}
Let \cstar{} be the wrong intermediate claim that sits inside the agent's own \role{<thought>}; it is given by construction, since the failure pool of Sec.~\ref{sec:eval-stats} injects a known wrong claim. We do not attempt to localize such errors automatically; that is a separate problem, treated in Appendix~\ref{app:localization}. The source-conditioned role-relabel intervention is the single-step operation of (i)~appending a message to the prompt whose payload is byte-identical to \cstar, (ii)~wrapping that payload in an external chat-template role (\role{user}, \role{tool}, or \role{system <memory>}), and (iii)~appending an audit instruction held byte-identical across all conditions. No tokens of \cstar{} change; only the role tags around the new copy change. Sec.~\ref{sec:byte-identity} formalizes the byte-identity guarantee.

The five conditions we compare are summarized in Table~\ref{tab:conditions}. The audit instruction is held byte-identical across all five; only the message role of \cstar{} varies. \cond{L0\_self} is the audit-only baseline: \cstar{} stays inside its original \role{<thought>}, and the agent is asked to audit. \cond{L\_user\_wait} and \cond{L\_user\_neutral} re-present \cstar{} as a \role{user} message, with and without the ``Wait,'' prefix, respectively; the contrast isolates the contribution of the prefix. \cond{L\_tool} re-presents \cstar{} as a calculator-tool response (name = \role{calculator}), exercising the \role{tool} role token of each model's chat template. \cond{L\_memory} re-presents \cstar{} inside a \role{system <memory>} block, the strongest source-of-record framing in the chat-template vocabulary.

\begin{table}[t]
\centering
\small
\caption{The five conditions compared in every experiment. The audit message is byte-identical across all five; only the wrapper of \cstar{} changes. \cond{L0\_self} omits the re-presented copy entirely and provides the audit-only baseline.}
\label{tab:conditions}
\setlength{\tabcolsep}{4pt}
\begin{tabular*}{\columnwidth}{@{\extracolsep{\fill}} l p{0.62\columnwidth} @{}}
\toprule
\textbf{Condition} & \textbf{Wrapper around \cstar} \\
\midrule
\cond{L0\_self}         & none (audit-only control) \\
\cond{L\_user\_wait}    & a \role{user} message: ``Wait, \cstar'' \\
\cond{L\_user\_neutral} & a \role{user} message containing only \cstar \\
\cond{L\_tool}          & a \role{tool} response containing \cstar \\
\cond{L\_memory}        & a \role{system} message with \role{<memory>}\cstar\role{</memory>} \\
\bottomrule
\end{tabular*}
\end{table}

\begin{table*}[t]
\centering
\small
\caption{Per-experiment paired increase in explicit-correction rate (CR) over the audit-only \cond{L0\_self} baseline, in pp. $L_0$ is the baseline CR; bold marks the strongest significant condition(s) per row, ties bolded jointly. $^{*}p<0.05$, $^{**}p<0.01$, $^{***}p<0.001$ (two-sided paired bootstrap); all three pre-specified criteria pass on the two 70B-class math experiments.}
\label{tab:final-main}
\begin{tabular*}{\textwidth}{@{\extracolsep{\fill}} l l l l l r @{}}
\toprule
\textbf{Model and domain ($n=30$)} & \cond{wait} & \cond{neutral} & \cond{tool} & \cond{memory} & $L_0$ \\
\midrule
\multicolumn{6}{l}{\textit{Open-weight, math}} \\
\quad Qwen2.5-72B math                  & $+23.3^{*}\phantom{^{**}}$ & $+26.7^{**}\phantom{^{*}}$ & $+26.7^{***}$ & $\mathbf{+53.3^{***}}$ & 16.7\% \\
\quad Llama-3.3-70B math                & $+56.7^{***}$ & $+60.0^{***}$ & $+3.3\phantom{^{***}}$ & $\mathbf{+86.7^{***}}$ & 0.0\%  \\
\quad gpt-oss-20B math (reasoning)      & $+6.7\phantom{^{***}}$ & $+16.7\phantom{^{***}}$ & $+3.3\phantom{^{***}}$ & $+16.7\phantom{^{***}}$ & 76.7\% \\
\midrule
\multicolumn{6}{l}{\textit{Cross-domain (logical deduction)}} \\
\quad Qwen2.5-72B gen-logic             & $+70.0^{***}$ & $\mathbf{+80.0^{***}}$ & $+56.7^{***}$ & $+63.3^{***}$ & 0.0\%  \\
\quad Llama-3.3-70B gen-logic           & $+60.0^{***}$ & $\mathbf{+93.3^{***}}$ & $+26.7^{***}$ & $+76.7^{***}$ & 0.0\%  \\
\quad Qwen2.5-72B BBH-LD                & $+10.0\phantom{^{***}}$ & $+6.7\phantom{^{***}}$ & $+16.7\phantom{^{***}}$ & $+13.3\phantom{^{***}}$ & 66.7\% \\
\midrule
\multicolumn{6}{l}{\textit{Closed-weight, math}} \\
\quad GPT-4o math                       & $+20.0\phantom{^{***}}$ & $\mathbf{+50.0^{***}}$ & $+46.7^{***}$ & $+43.3^{***}$ & 40.0\% \\
\quad Claude Sonnet 4 math              & $\mathbf{+40.0^{***}}$ & $\mathbf{+40.0^{***}}$ & $+36.7^{***}$ & $\mathbf{+40.0^{***}}$ & 53.3\% \\
\quad Gemini 2.5 Flash-Lite math        & $+6.7\phantom{^{***}}$ & $+6.7\phantom{^{***}}$ & $0.0\phantom{^{***}}$ & $\mathbf{+23.3^{*}}\phantom{^{**}}$ & 66.7\% \\
\midrule
\multicolumn{6}{l}{\textit{Additional open-weight families}} \\
\quad Gemma-3-12B math                  & $+63.3^{***}$ & $\mathbf{+66.7^{***}}$ & $0.0\phantom{^{***}}$ & $\mathbf{+66.7^{***}}$ & 30.0\% \\
\quad Phi-4-14B math                    & $+66.7^{***}$ & $+73.3^{***}$ & $+73.3^{***}$ & $\mathbf{+80.0^{***}}$ & 20.0\% \\
\quad Qwen3-30B math (reasoning)        & $+3.3\phantom{^{***}}$ & $-3.3\phantom{^{***}}$ & $0.0\phantom{^{***}}$ & $\mathbf{+30.0^{**}}\phantom{^{*}}$ & 53.3\% \\
\bottomrule
\end{tabular*}
\end{table*}

\subsection{Byte-Identity}
\label{sec:byte-identity}

The intervention swaps the chat-template role around \cstar{} while leaving the wrapped text untouched. To make this guarantee operational, we assert at every trial that the injected \cstar{} message payload is byte-identical to the original by a SHA-256 (Secure Hash Algorithm 256-bit) digest match, so no tokens of \cstar{} change across conditions; the wrapping role and each template's boundary characters (for example \role{<|im\_start|>system\textbackslash n}) are supplied by the model's native chat template (Appendix~\ref{app:chat-templates}). Trials failing the hash check are discarded and re-generated. Byte-identity lets any difference in correction rate across conditions be attributed to the role tag, not to a change in the claim's wording.

\subsection{Separating Wrapper and Role Effects}
\label{sec:handle-granularity}

The intervention introduces two ingredients at once: a syntactic wrapper around \cstar, and a specific role tag attached to that wrapper. To separate their contributions, we construct a sequence of conditions that progressively varies the wrapper while holding everything else fixed. The ladder runs from no wrapper (\cond{H0}~$=$~\cond{L0\_self}) through label-free boundary marks (\cond{H1} bare brackets, \cond{H2} explicit prefix tag, \cond{H3} XML-style wrapper with no role) up to the full role-tagged condition (\cond{H4}~$=$~\cond{L\_memory}). The \cond{H1}--\cond{H3} rungs bound the contribution of a bare syntactic boundary, and the \cond{H3}--\cond{H4} contrast isolates the additional contribution of the role tag itself. Spanning \cond{H1}--\cond{H3} across a bracket, a prefix tag, and an XML wrapper guards against any single surface form driving the effect. Applied to the Qwen-72B math failure pool (Sec.~\ref{sec:exp-overall}, Table~\ref{tab:handle}), this ladder attributes the increase to two additive parts, a label-free boundary and the role tag on top of it.

%%%%%%%%%%%%%%%%%%%%%%%%%%%%%%%%%%%%%%%%%%%%%%%%%%%%%%%%%%%%%%%%%%%%%%%
%%%%%%%%%%%%%%%%%%%%%%%%%%%%% Experiment %%%%%%%%%%%%%%%%%%%%%%%%%%%%%%
%%%%%%%%%%%%%%%%%%%%%%%%%%%%%%%%%%%%%%%%%%%%%%%%%%%%%%%%%%%%%%%%%%%%%%%
\section{Experiments}
\label{sec:experiments}

\subsection{Experimental Setting}
\label{sec:exp-setting}

We evaluated nine models in three domains. The open-weight set was served via Ollama and comprised two 70B-class models, Qwen2.5-72B-Instruct (Qwen-72B)~\citep{qwenQwen25TechnicalReport2025} and Llama-3.3-70B-Instruct (Llama-70B)~\citep{dubeyLlama3Herd2024}, the reasoning-tuned gpt-oss-20B~\citep{openaiGptOssModelCard2025}, plus three smaller families for breadth, Gemma-3-12B~\citep{teamGemma3Technical2025}, Phi-4-14B~\citep{abdinPhi4TechnicalReport2024a}, and Qwen3-30B~\citep{qwenQwen3TechnicalReport2025} with reasoning enabled; three closed-weight frontier models were served via APIs, GPT-4o~\citep{openaiGPT4oSystemCard2024}, Claude Sonnet 4~\citep{anthropicClaude4SystemCard2025}, and Gemini 2.5 Flash-Lite~\citep{comaniciGemini25Pushing2025a}. Because these families realize the \role{tool} role with materially different template tokens, the \cond{L\_tool} condition doubled as a test of whether a distinctive tool-role token is needed for the increase in correction rate (Sec.~\ref{sec:exp-overall} shows it is not needed). One family, Gemma-3, has no native \role{tool} role at all, so for it the \cond{L\_tool} condition could not be realized and reduced to the audit-only baseline (Appendix~\ref{app:chat-templates}); we therefore treated its \cond{L\_tool} result as not applicable rather than as a null role effect. All three domains used a chain-of-thought formulation~\citep{weiChainofThoughtPromptingElicits2023,kojimaLargeLanguageModels2023}: \textbf{Math} (GSM8K-style arithmetic, where \cstar{} is a wrong arithmetic value, e.g.\ $5\times16=96$ when the correct value is $80$), \textbf{Generated logic} (synthesized transitive-ordering puzzles over five entities, where \cstar{} is a wrong non-adjacent ordering only multi-step reasoning can check, e.g.\ claiming one entity outranks another against the transitive chain), and \textbf{BBH-LD} (the Logical Deduction five-objects subtask of BBH~\citep{suzgunChallengingBIGBenchTasks2022,srivastavaImitationGameQuantifying2023}, where \cstar{} names a wrong option as the established intermediate inference). Of the three domains, only generated logic admits no cheap verification: a transitive-ordering claim cannot be verified by a quick recalculation, but requires full multi-step reasoning. This makes it the sharpest test that the effect is not a mere verification cue (Sec.~\ref{sec:exp-generalization}).

\subsection{Failure Pool and Scoring}
\label{sec:eval-stats}
We evaluated on a failure pool where intrinsic self-correction fails without an audit instruction, concentrating statistical power on the target regime and removing baseline-difficulty confounds. For each task we obtained the agent's clean reasoning trajectory $\mathcal{T}_0$, injected \cstar{} as a fresh \role{<thought>} block to form the perturbed trajectory $\mathcal{T}_1$, resumed two turns without the audit instruction, and retained tasks whose continuations were scored uncorrected. The \cond{L0\_self} baseline then appends the audit, so its CR is not zero by construction. Each retained task then ran under the five conditions of Table~\ref{tab:conditions} and, on Qwen-72B math, the \cond{H0}--\cond{H4} ladder of Sec.~\ref{sec:handle-granularity}. Scoring used a locked Qwen-72B judge, i.e.\ a fixed LLM-as-judge with frozen model, prompt, and sampling temperature $T=0$, that returns \texttt{YES}/\texttt{NO} on whether the continuation explicitly identifies \cstar{} as wrong, following the calibration guidance of~\citet{kimChallengingEvaluatorLLM2025}; the explicit-correction rate (CR) is the \texttt{YES} fraction, with Cohen's $\kappa=1.0$ against a hand-labeled set of $N$ trials and $\kappa=0.843$ under an independent re-judge (Appendix~\ref{app:stats-multijudge}). Each experiment drew $n=30$ paired tasks at $T=0$ with a fixed seed; we reported $\Delta\text{CR}$ against \cond{L0\_self} with two-sided $p$-values and 95\% CIs from a 10{,}000-sample paired bootstrap ($^{*}p<0.05$, $^{**}p<0.01$, $^{***}p<0.001$). Three success criteria were pre-specified on the two 70B-class math experiments: (i) on Qwen-72B, the strongest condition raises CR by $\geq 20$\,pp at $p<0.01$; (ii) \cond{L\_user\_neutral} exceeds $+5$\,pp there, separating the role handle from the ``Wait,'' trigger; and (iii) Llama-70B replicates in the same direction by $\geq 15$\,pp.

\subsection{Main Results}
\label{sec:exp-overall}

Table~\ref{tab:final-main} reports all 12 model-domain combinations, and all three pre-specified criteria (Sec.~\ref{sec:eval-stats}) passed on the two 70B-class math experiments. A significant increase appeared on at least one relabel condition in 10 of the 12 combinations, and per-task flips are directly causal: holding the trajectory and byte-identical \cstar{} fixed, at least one relabel flipped a response to corrected in 29 of 30 eligible (baseline-uncorrected) tasks on Llama-70B logic and 18 of 25 on Qwen-72B math. The two exceptions (gpt-oss-20B math, Qwen-72B BBH-LD) are predicted: a high $L_0$ leaves little headroom, and BBH-LD additionally permits option elimination. The effect survived Holm-Bonferroni correction (9 of 12 combinations) and a second judge ($\kappa=0.843$; Appendices~\ref{app:stats-multicomp},~\ref{app:stats-multijudge}).

We also found that the most effective wrapper is model- and domain-dependent. \cond{L\_memory} led in 7 combinations and \cond{L\_user\_neutral} in 5. \cond{L\_memory} dominated on most math experiments, consistent with arithmetic claims being out-of-genre for a user turn, whereas \cond{L\_user\_neutral} was strongest on logical deduction, a natural user-style genre. The frontier replication held on GPT-4o and Claude Sonnet 4 (unsaturated baselines), whereas Gemini 2.5 Flash-Lite showed only one significant lift (\cond{L\_memory}, $+23.3$\,pp), consistent with its higher $L_0$ of $66.7\%$. The gain is not merely a verification cue: on generated logic, whose transitive-ordering errors cannot be caught by quick recomputation, both 70B open-weight models increased significantly (up to $+93.3$\,pp on Llama-70B), whereas BBH-LD, whose multiple-choice format already permits option-elimination, showed only smaller non-significant gains above its $66.7\%$ baseline (largest $+16.7$\,pp, $p=0.06$).

Because every relabel condition re-presents \cstar{} explicitly, one might worry the gain is merely a localization hint; two controls ruled this out. A duplication control that re-presents \cstar{} at the same position but still inside \role{<thought>} barely helped ($+6.7$\,pp, $p=0.26$), whereas the matched \role{<memory>} relabel reached a $+46.7$\,pp role effect (Appendix~\ref{app:mechanism-recency}); the wrapper-role ladder (Table~\ref{tab:handle}) adds that the role tag contributes a further $30$\,pp beyond a label-free wrapper. At fixed \role{system} role, a nonsense \role{<xqzy>} still trailed \role{<memory>}, so the tag's lexical identity also mattered (Appendix~\ref{app:mechanism-tokens}). A DeepSeek-R1 check (57 trials, $n=10$--$13$ per condition) confirmed the saturated-baseline case: every condition, including \cond{L0\_self}, reached $100\%$ (Appendix~\ref{app:reasoning}).

\begin{table}[t]
\centering
\small
\setlength{\tabcolsep}{4pt}
\caption{Wrapper-role ladder on the Qwen-72B math failure pool ($n=30$). \cond{H0} omits the wrapper; \cond{H1}--\cond{H3} introduce label-free boundaries; \cond{H4} adds the \role{system} role tag. $\Delta$ is the increase over \cond{H0}, in pp.}
\label{tab:handle}
\begin{tabular*}{\columnwidth}{@{\extracolsep{\fill}} p{0.65\columnwidth} r r @{}}
\toprule
\textbf{Rung (wrapper of \cstar)} & CR & $\Delta$ \\
\midrule
\cond{H0} no wrapper (= \cond{L0\_self}) & 16.7\% & --- \\
\cond{H1} bare brackets \texttt{[\cstar]} & 33.3\% & $+16.7$ \\
\cond{H2} explicit prefix \texttt{NOTE: \cstar} & 36.7\% & $+20.0$ \\
\cond{H3} XML wrapper \texttt{<x>\cstar</x>} (no role) & 40.0\% & $+23.3$ \\
\cond{H4} \role{system} \texttt{<memory>\cstar</memory>} (= \cond{L\_memory}) & \textbf{70.0\%} & $\mathbf{+53.3}$ \\
\bottomrule
\end{tabular*}
\end{table}

\subsection{From Verification to Addressability}
\label{sec:exp-addressability}
To show that the source-conditioned role relabeling does not trigger self-correction just by cueing verification, we conducted an analysis on self-directed verification. In the Qwen-72B math failure pool, we replaced the audit instruction with three self-distrust prompt variants while leaving \cstar{} inside the agent's original \role{<thought>} (Table~\ref{tab:decisive-control}). None approached the $70\%$ of \cond{L\_memory}: the strongest reached $23.3\%$ and two worsened the baseline, since a blanket distrust cue scatters attention across all intermediates rather than onto \cstar. For all three prompt variants, the agent recomputed the answer correctly but rarely named or rejected the specific wrong intermediate, so the judge usually returned \texttt{NO}. The relabel instead let the agent produce phrases such as \textit{``the memory says X; that is incorrect because $\ldots$''}. Additionally, re-scoring final answer only yielded marginal changes in correction rates ($\pm 17$\,pp, $p>0.05$; Appendix~\ref{app:final-answer-metric}). Established methods that add reasoning passes rather than a discrete referent (Self-Refine, Reflexion, and Chain-of-Verification) never matched the relabel on any unsaturated baseline, and on the open-weight pools reversed the increase under the strict criterion (Appendix~\ref{app:baselines}). A first-token log-probability analysis on Qwen-72B corroborated this naming behavior at the token level: relabeling \cstar{} into \role{<memory>} raised the probability of an opening ``engagement'' token that names the source before rejecting it (e.g.\ ``Indeed''), a lexical signature that was model-specific and did not reproduce across the other families, even though the behavioral increase did (Appendix~\ref{app:mechanism-tokens-models}). Collectively, these findings suggest that the primary cause behind self-correction failures is not verification capability but \emph{addressability}: the ability to treat a specific claim as a discrete, nameable object the agent can act on.

\begin{table}[t]
\centering
\small
\setlength{\tabcolsep}{4pt}
\caption{Self-distrust prompt variants on the Qwen-72B math failure pool ($n=30$). The trailing audit is varied while \cstar{} stays inside its original \role{<thought>}. $\Delta$ values are in pp.}
\label{tab:decisive-control}
\begin{tabular*}{\columnwidth}{@{\extracolsep{\fill}} @{} p{0.60\columnwidth} r r @{}}
\toprule
\textbf{Appended user prompt} & \textbf{CR} & \shortstack[r]{\textbf{$\Delta$ to}\\\textbf{\cond{L\_memory}}} \\
\midrule
``Audit your reasoning'' (\cond{L0\_self}) & 16.7\% & $-53.3$ \\
\midrule
``Previous thoughts may contain errors'' & 3.3\% & $-66.7$ \\
``Intermediate arithmetic may be wrong'' & 0.0\% & $-70.0$ \\
``Restate intermediates, then verify'' & 23.3\% & $-46.7$ \\
\midrule
(reference) \cond{L\_memory} & \textbf{70.0\%} & --- \\
\bottomrule
\end{tabular*}
\end{table}

\subsection{Generalization, Robustness, and Ablations}
\label{sec:exp-generalization}
\label{sec:exp-robustness}

\begin{table}[t]
\centering
\small
\caption{Variance check at $T=0.5$, three seeds, 15-task Qwen subset. s0--s2 are per-seed correction rates; Agg.\ $\Delta$ is the aggregate increase over \cond{L0\_self} (pp).}
\label{tab:variance}
\begin{tabular*}{\columnwidth}{@{\extracolsep{\fill}}lcccrc@{}}
\toprule
\textbf{Condition} & s0 & s1 & s2 & Agg.\ $\Delta$ & $p$ \\
\midrule
\cond{L0\_self}         & 13\% & 27\% & 33\% & ---   & --- \\
\cond{L\_user\_wait}    & 20\% & 40\% & 33\% & $+7$  & $0.23\phantom{^{**}}$ \\
\cond{L\_user\_neutral} & 53\% & 40\% & 33\% & $+18$ & $0.04^{*}\phantom{^{*}}$ \\
\cond{L\_tool}          & 33\% & 33\% & 33\% & $+9$  & $0.18\phantom{^{**}}$ \\
\cond{L\_memory}        & 33\% & 67\% & 60\% & $\mathbf{+29}$ & $\mathbf{0.001^{**}}$ \\
\bottomrule
\end{tabular*}
\end{table}

\begin{table}[t]
\centering
\small
\caption{Commitment-depth check on Llama-70B math ($n=30$). $\Delta$ is the paired increase over \cond{L0\_self} (pp) at the just-injected state (Shallow) versus one turn later (Deep).}
\label{tab:depth}
\begin{tabular*}{\columnwidth}{@{\extracolsep{\fill}} l r r r @{}}
\toprule
\textbf{Condition} & Shallow $\Delta$ & Deep $\Delta$ & Change \\
\midrule
\cond{L\_user\_wait}    & $+57^{***}$ & $+63^{***}$ & $+6$ \\
\cond{L\_user\_neutral} & $+60^{***}$ & $+70^{***}$ & $+10$ \\
\cond{L\_tool}          & $+3\phantom{^{***}}$ & $+7\phantom{^{***}}$ & $+4$ \\
\cond{L\_memory}        & $+87^{***}$ & $\mathbf{+93^{***}}$ & $+6$ \\
\bottomrule
\end{tabular*}
\end{table}

Two analyses further confirm the increase is not a deterministic-setting artifact. At $T=0.5$ over three seeds (15-task Qwen subset), \cond{L\_memory}'s increase stayed significant ($p<0.01$, Table~\ref{tab:variance}), and a five-seed re-sampling of \cond{L0\_self} held at $16.0\%$ versus the main-run $16.7\%$ (Appendix~\ref{app:stats-stochasticity}). Nor did the effect decay with \emph{commitment depth}: relabeling one turn later (Deep) rather than at injection (Shallow) left every increase intact or slightly larger (Table~\ref{tab:depth}), since the relabel adds a fresh external referent rather than overwriting already-committed tokens.

A third check addressed audit wording, testing whether the increase is specific to one fixed phrasing rather than to the role label. Holding the Qwen-72B math failure pool, \cstar{}, and the \role{<memory>} wrapper fixed, we varied only the audit across five lexically disjoint paraphrases:
\begin{itemize}\small
\item[\textbf{V1}] (canonical) ``Audit your reasoning so far. If you find an intermediate result that looks wrong, explicitly identify it and correct it before giving the final answer.''
\item[\textbf{V2}] ``Review your previous steps. Point out any wrong intermediate value by name and fix it, then state the final answer.''
\item[\textbf{V3}] ``Check each intermediate calculation. If one is incorrect, say which one and correct it before answering.''
\item[\textbf{V4}] ``Go back over your work. Name any mistake in the intermediate values, revise it, and then give the final answer.''
\item[\textbf{V5}] ``Examine your prior reasoning. Call out any incorrect intermediate result and replace it with the correct value before concluding.''
\end{itemize}

The paired \cond{L\_memory}-over-\cond{L0\_self} increase was positive under all five paraphrases (Table~\ref{tab:audit-paraphrase}) and significant in four ($\Delta \in [+26.7, +53.3]$\,pp), indicating the effect is robust to audit wording. The one marginal case, V2 ($+6.7$\,pp), is also the only paraphrase whose audit itself names an explicit-address cue (``point out any wrong intermediate value by name''); this may raise its \cond{L0\_self} baseline ($33.3\%$) and leave less headroom, but we do not draw a strong conclusion from a single case.

\begin{table}[t]
\centering
\small
\caption{Audit-wording robustness on the Qwen-72B math failure pool ($n=30$). $\Delta$ is the paired increase of \cond{L\_memory} over \cond{L0\_self}.}
\label{tab:audit-paraphrase}
\begin{tabular*}{\columnwidth}{@{\extracolsep{\fill}} l r r r @{}}
\toprule
\textbf{Audit variant} & $L_0$ & $L_{\text{mem.}}$ & $\Delta$ \\
\midrule
V1 (canonical)        & 16.7\% & \textbf{70.0\%} & $\mathbf{+53.3^{***}}$       \\
V2 (review/point out) & 33.3\% & 40.0\%          & $+6.7\phantom{^{***}}$      \\
V3 (check each)       & 10.0\% & 50.0\%          & $+40.0^{***}$               \\
V4 (go back over)     & 26.7\% & 53.3\%          & $+26.7^{*}\phantom{^{**}}$  \\
V5 (examine prior)    & 16.7\% & 46.7\%          & $+30.0^{*}\phantom{^{**}}$  \\
\bottomrule
\end{tabular*}
\end{table}

\subsection{Adversarial Mirror and Safety Scope}
\label{sec:exp-adversarial}

Is \role{<memory>}, the strongest lever for making the agent \emph{reject} a wrong claim, also the strongest for making it \emph{accept} one? To answer this question, we conducted adversarial attacks on tasks the agent had solved correctly, injecting \cstar{} into each external role. An attack is counted as successful if the final answer adopts \cstar. Across five experiments including two models, two domains, and calculator-available versus no-tools regimes, adversarial injection resulted in similar or lower error rates compared to baseline, irrespective of the injection target (Table~\ref{tab:adversarial}). The effect did not reverse: under the neutral adversarial prompt, and on tasks selected for correct baseline solutions, the agent adopted a wrong claim from its own \role{<thought>} about $83\%$ of the time on Qwen-72B math, but a byte-identical claim under an external role at most $3.3\%$, the same self-versus-external asymmetry in the reverse direction. Unlike indirect prompt injection~\citep{greshakeNotWhatYouve2023,bagdasaryanAbusingImagesSounds2023}, here the untrusted content that drives errors is the agent's own intermediate, not external input. This low attack-success rate held by default but could be overridden. Holding the \role{<memory>} wrapper fixed and varying only the trailing audit, a single sentence ordering the agent to treat the memory as truth and not verify raised the attack rate from $3.3\%$ to $70\%$, while a distrust framing left it unchanged (Table~\ref{tab:instructibility}). Verify-by-default is therefore a learned response to the prompt context, not an architectural guarantee, which is why the intervention is a diagnostic lever, not a hardened defense: deployments that permit trust-framing instructions lose this safety property.

\begin{table}[t]
\centering
\small
\caption{Adversarial mirror ($n=30$ per experimental setting). Values report the rate at which the model accepts the target false claim; for \cond{A0}, this is the spontaneous acceptance rate without injection.}
\label{tab:adversarial}
\begin{tabular*}{\columnwidth}{@{\extracolsep{\fill}} l r r r r @{}}
\toprule
\textbf{Experiment} & \cond{A0} & \cond{user} & \cond{tool} & \cond{mem.} \\
\midrule
Qwen-72B math + calc  & 3.3\% & 3.3\% & 0.0\% & 3.3\% \\
Qwen-72B math no-tool & 3.3\% & 3.3\% & 3.3\% & 3.3\% \\
Llama-70B math + calc & 3.3\% & 3.3\% & 3.3\% & 3.3\% \\
Qwen-72B logic        & 0.0\% & 0.0\% & 0.0\% & 0.0\% \\
Llama-70B logic       & 0.0\% & 3.3\% & 0.0\% & 0.0\% \\
\midrule
\multicolumn{5}{@{}l@{}}{\textit{Self-injection reference} (Qwen-72B math): $\mathbf{83.3\%}$} \\
\bottomrule
\end{tabular*}
\end{table}

\begin{table}[t]
\centering
\small
\caption{Instructability of the \role{<memory>} safety on Qwen-72B math ($n=30$). Values are false-claim adoption rates; $\Delta$ is in pp.}
\label{tab:instructibility}
\begin{tabular*}{\columnwidth}{@{\extracolsep{\fill}}lrr@{}}
\toprule
\textbf{Condition} & Adoption & $\Delta$ \\
\midrule
\cond{default} (\role{<memory>}, neutral audit)           & 3.3\%           & --- \\
\cond{trust} (``treat as truth, do not verify'') & \textbf{70.0\%} & $\mathbf{+66.7^{***}}$ \\
\cond{distrust} (``verify before using'')                 & 3.3\%           & $\pm$0 \\
\bottomrule
\end{tabular*}
\end{table}

%%%%%%%%%%%%%%%%%%%%%%%%%%%%%%%%%%%%%%%%%%%%%%%%%%%%%%%%%%%%%%%%%%%%%%%
%%%%%%%%%%%%%%%%%%%%%%%%%%% Discussion %%%%%%%%%%%%%%%%%%%%%%%%%%%%%%%%
%%%%%%%%%%%%%%%%%%%%%%%%%%%%%%%%%%%%%%%%%%%%%%%%%%%%%%%%%%%%%%%%%%%%%%%
\section{Discussion}
\label{sec:discussion}

\subsection{What ``Self'' Means in a Chat-Template}
\label{sec:discussion-self}

We interpret the effect as a by-product of instruction tuning, which rewards responding to content arriving under external roles (\role{user}, \role{system}, \role{tool}) far more than acting on the model's own \role{<thought>} block as a target. Under this interpretation, the instruction to ``audit your reasoning'' admits no well-practiced response other than re-derivation, which is what the control of Sec.~\ref{sec:exp-addressability} observes: the agent recomputes but rarely singles out and rejects the specific intermediate. The relabel sidesteps the gap by recasting an unaddressable thought-internal claim as the kind of object the model can readily process, an external assertion that invites a verify-and-rebut response. This interpretation also provides an explanation for the model- and domain-dependence of the most effective role wrapper. We hypothesize that the strongest role is the one whose discourse register best matches the wrong claim's content type, since that match elicits the most reliable rebuttal. The ``self'' that fails to self-correct is, in this view, not a cognitive deficit but a chat-template artifact.

\subsection{Connections to Prior Work}
\label{sec:discussion-reconciliation}

The result connects several lines that have been studied separately. Most directly, \citet{tsuiSelfCorrectionBench2025} establishes the blind spot and reads ``Wait'' as a correction trigger; we show the trigger is incidental and the carrier role primary, recasting the gap as a failure of addressability rather than a missing verification step. The unfaithful-CoT literature~\citep{turpinLanguageModelsDont2023,lanhamMeasuringFaithfulnessChainofThought2023,chenReasoningModelsDont2025} shows a reasoning trace can diverge from the computation behind the answer. We add its corrective face, showing the trace is also not addressable as a discrete object even when its content remains revisable. The user-assistant bias of~\citet{panUserAssistantBiasLLMs2026} captures the same role-asymmetry in static completion, which we carry into the corrective regime as a diagnostic lever, not only a vulnerability. The memory-poisoning literature~\citep{dongMemoryInjectionAttacks2026} reports high attack success from narrative-coherent payloads. Our null adversarial result occupies the complementary regime of bare injection, and the instructability experiment shows trust framing, not the role tag, is the gateway. Finally,~\citet{wallaceInstructionHierarchyTraining2024} predict \role{system} outranks \role{user} content; we show this is overridable from the user role, recasting role priority as behavioral, not architectural.

\subsection{Scope and Limitations}
\label{sec:discussion-limitations}
Self-correction improves most for models with unsaturated audit-only baselines, chiefly standard instruction-tuned ones. Models whose baseline is already saturated leave less headroom (Appendices~\ref{app:reasoning},~\ref{app:baselines}), consistent with the addressability account: models that already identify errors under audit-only gain little from an external referent. Although $n=30$ per setting supports the effects, it limits analyses by claim length, audit distance, and number of competing intermediates. The strict-identification criterion restricts the study to verifiable tasks; extension to free-form reasoning, code debugging, and planning remains future work. Deployment assumes that \cstar{} has been localized, although identifying the erroneous step remains an unresolved problem~\citep{tyenLLMsCannotFind2024}. We therefore view the intervention as a mechanistic probe and a relabeling stage dependent on an upstream detector (Appendix~\ref{app:localization}). Safety is conditional: a trust-framing prompt raises the attack rate to $70\%$ (Sec.~\ref{sec:exp-adversarial}), making the intervention a diagnostic lever rather than a hardened defense. Behavioral controls support the addressability account: self-distrust prompting and within-trace duplication produce smaller gains than presenting the same claim as externally authored. However, a final-layer hidden-state probe does not predict trial-level correction above baseline, leaving the internal mechanism unresolved (Appendix~\ref{app:mechanism-probe}). Three qualifications remain. First, the failure pool includes only tasks on which audit-only fails strict identification, so the gains characterize this targeted regime rather than in-the-wild prevalence. Second, \cond{L\_tool} realization varies across chat templates, and one family (Gemma-3) has no native \role{tool} role, so part of the cross-family \cond{L\_tool} variance reflects template limitations rather than role effects (Appendix~\ref{app:chat-templates}). Third, the GPT-4o and GPT-4.1 replications use fewer than 30 tasks due to rate limits, resulting in wider CIs and suggestive comparisons (Appendix~\ref{app:baselines}). These limitations narrow the claims but not the central behavioral finding.
%%%%%%%%%%%%%%%%%%%%%%%%%%%%%%%%%%%%%%%%%%%%%%%%%%%%%%%%%%%%%%%%%%%%%%%
%%%%%%%%%%%%%%%%%%%%%%%%%%% Conclusion %%%%%%%%%%%%%%%%%%%%%%%%%%%%%%%%
%%%%%%%%%%%%%%%%%%%%%%%%%%%%%%%%%%%%%%%%%%%%%%%%%%%%%%%%%%%%%%%%%%%%%%%

\section{Conclusion}
\label{sec:conclusion}

An LLM agent's failure to self-correct a wrong claim inside its own \role{<thought>} is, to a large degree, not a reasoning deficit but a chat-template artifact. Re-presenting the byte-identical claim under an external role consistently increases the explicit-correction rate, the rate at which the agent names and rejects the claim, across the model families and domains we study, whereas a self-distrust prompt that leaves the claim in place does not. Our experiments isolate the mechanism as addressability rather than purely verification: the agent retains the capability to check the claim, and often re-derives the right answer silently, but has no learned way to act on a thought-internal substring as a discrete object; the relabel supplies the missing handle. These findings turn a studied limitation into a controllable, zero-training lever for explicitly surfacing self-generated errors, bounded by an asymmetric safety that a single trust-framing sentence can override. Whether addressability also governs self-correction in free-form reasoning, beyond the verifiable tasks studied here, is the natural next question, one we expect the same role-handle to illuminate.

% References
\bibliography{aaai2027}

@book{abdinPhi4TechnicalReport2024a,
  title = {Phi-4 {{Technical Report}}},
  author = {Abdin, Marah and Aneja, Jyoti and Behl, Harkirat and Bubeck, S{\'e}bastien and Eldan, Ronen and Gunasekar, Suriya and Harrison, Michael and Hewett, Russell and Javaheripi, Mojan and Kauffmann, Piero and Lee, James and Lee, Yin Tat and Li, Yuanzhi and Liu, Weishung and Mendes, Caio and Nguyen, Anh and Price, Eric and {de Rosa}, Gustavo and Saarikivi, Olli and Zhang, Yi},
  year = 2024,
  month = dec,
  doi = {10.48550/arXiv.2412.08905}
}

@misc{anthropicClaude4SystemCard2025,
  title = {{{System Card}}: {{Claude Opus}} 4 \& {{Claude Sonnet}} 4},
  shorttitle = {Claude 4 {{System Card}}},
  author = {{Anthropic}},
  year = 2025,
  month = may,
  publisher = {Anthropic},
  urldate = {2026-05-25},
  howpublished = {https://www-cdn.anthropic.com/4263b940cabb546aa0e3283f35b686f4f3b2ff47.pdf},
  langid = {english}
}

@misc{bagdasaryanAbusingImagesSounds2023,
  title = {Abusing {{Images}} and {{Sounds}} for {{Indirect Instruction Injection}} in {{Multi-Modal LLMs}}},
  author = {Bagdasaryan, Eugene and Hsieh, Tsung-Yin and Nassi, Ben and Shmatikov, Vitaly},
  year = 2023,
  month = oct,
  number = {arXiv:2307.10490},
  eprint = {2307.10490},
  primaryclass = {cs.CR},
  publisher = {arXiv},
  doi = {10.48550/arXiv.2307.10490},
  urldate = {2026-05-25},
  archiveprefix = {arXiv}
}

@misc{bhatiaDistributionalSemanticsTracing2026,
  title = {Distributional {{Semantics Tracing}}: {{A Framework}} for {{Explaining Hallucinations}} in {{Large Language Models}}},
  shorttitle = {Distributional {{Semantics Tracing}}},
  author = {Bhatia, Gagan and Sripada, Somayajulu G. and Allan, Kevin and Azcona, Jacobo},
  year = 2026,
  month = mar,
  number = {arXiv:2510.06107},
  eprint = {2510.06107},
  primaryclass = {cs.CL},
  publisher = {arXiv},
  doi = {10.48550/arXiv.2510.06107},
  urldate = {2026-05-25},
  archiveprefix = {arXiv}
}

@inproceedings{chenEnigmataScalingLogical2025b,
  title = {Enigmata: {{Scaling Logical Reasoning}} in {{Large Language Models}} with {{Synthetic Verifiable Puzzles}}},
  shorttitle = {Enigmata},
  booktitle = {The {{Thirty-ninth Annual Conference}} on {{Neural Information Processing Systems}}},
  author = {Chen, Jiangjie and He, Qianyu and Yuan, Siyu and Chen, Aili and Cai, Zhicheng and Dai, Weinan and Yu, Hongli and Chen, Jiaze and Li, Xuefeng and Yu, Qiying and Zhou, Hao and Wang, Mingxuan},
  year = 2025,
  month = oct,
  urldate = {2026-05-25},
  langid = {english}
}

@misc{chenReasoningModelsDont2025,
  title = {Reasoning {{Models Don}}'t {{Always Say What They Think}}},
  author = {Chen, Yanda and Benton, Joe and Radhakrishnan, Ansh and Uesato, Jonathan and Denison, Carson and Schulman, John and Somani, Arushi and Hase, Peter and Wagner, Misha and Roger, Fabien and Mikulik, Vlad and Bowman, Samuel R. and Leike, Jan and Kaplan, Jared and Perez, Ethan},
  year = 2025,
  month = may,
  number = {arXiv:2505.05410},
  eprint = {2505.05410},
  primaryclass = {cs.CL},
  publisher = {arXiv},
  doi = {10.48550/arXiv.2505.05410},
  urldate = {2026-05-25},
  archiveprefix = {arXiv}
}

@misc{chenTeachingLargeLanguage2023a,
  title = {Teaching {{Large Language Models}} to {{Self-Debug}}},
  author = {Chen, Xinyun and Lin, Maxwell and Sch{\"a}rli, Nathanael and Zhou, Denny},
  year = 2023,
  month = oct,
  number = {arXiv:2304.05128},
  eprint = {2304.05128},
  primaryclass = {cs.CL},
  publisher = {arXiv},
  doi = {10.48550/arXiv.2304.05128},
  urldate = {2026-05-25},
  archiveprefix = {arXiv}
}

@article{cobbeTrainingVerifiersSolve2021,
  title = {Training {{Verifiers}} to {{Solve Math Word Problems}}},
  author = {Cobbe, K. and Kosaraju, Vineet and Bavarian, Mo and Chen, Mark and Jun, Heewoo and Kaiser, Lukasz and Plappert, Matthias and Tworek, Jerry and Hilton, Jacob and Nakano, Reiichiro and Hesse, Christopher and Schulman, John},
  year = 2021,
  month = oct,
  journal = {ArXiv},
  urldate = {2026-05-27}
}

@book{comaniciGemini25Pushing2025a,
  title = {Gemini 2.5: {{Pushing}} the {{Frontier}} with {{Advanced Reasoning}}, {{Multimodality}}, {{Long Context}}, and {{Next Generation Agentic Capabilities}}},
  shorttitle = {Gemini 2.5},
  author = {Comanici, Gheorghe and Bieber, Eric and Schaekermann, Mike and Pasupat, Ice and Sachdeva, Noveen and Dhillon, Inderjit and Blistein, Marcel and Ram, Ori and Zhang, Dan and Rosen, Evan and Marris, Luke and Petulla, Sam and Gaffney, Colin and Aharoni, Asaf and Lintz, Nathan and Pais, Tiago and Jacobsson, Henrik and Szpektor, Idan and Jiang, Nan-Jiang and Hahn, Chris},
  year = 2025,
  month = jul,
  doi = {10.48550/arXiv.2507.06261}
}

@misc{dongMemoryInjectionAttacks2026,
  title = {Memory {{Injection Attacks}} on {{LLM Agents}} via {{Query-Only Interaction}}},
  author = {Dong, Shen and Xu, Shaochen and He, Pengfei and Li, Yige and Tang, Jiliang and Liu, Tianming and Liu, Hui and Xiang, Zhen},
  year = 2026,
  month = feb,
  number = {arXiv:2503.03704},
  eprint = {2503.03704},
  primaryclass = {cs.LG},
  publisher = {arXiv},
  doi = {10.48550/arXiv.2503.03704},
  urldate = {2026-05-25},
  archiveprefix = {arXiv}
}

@book{dubeyLlama3Herd2024,
  title = {The {{Llama}} 3 {{Herd}} of {{Models}}},
  author = {Dubey, Abhimanyu and Jauhri, Abhinav and Pandey, Abhinav and Kadian, Abhishek and {Al-Dahle}, Ahmad and Letman, Aiesha and Mathur, Akhil and Schelten, Alan and Yang, Amy and Fan, Angela and Goyal, Anirudh and Hartshorn, Anthony and Yang, Aobo and Mitra, Archi and Sravankumar, Archie and Korenev, Artem and Hinsvark, Arthur and Rao, Arun and Zhang, Aston and Zhao, Zhiwei},
  year = 2024,
  month = jul,
  doi = {10.48550/arXiv.2407.21783}
}

@misc{gaoPALProgramaidedLanguage2023,
  title = {{{PAL}}: {{Program-aided Language Models}}},
  shorttitle = {{{PAL}}},
  author = {Gao, Luyu and Madaan, Aman and Zhou, Shuyan and Alon, Uri and Liu, Pengfei and Yang, Yiming and Callan, Jamie and Neubig, Graham},
  year = 2023,
  month = jan,
  number = {arXiv:2211.10435},
  eprint = {2211.10435},
  primaryclass = {cs.CL},
  publisher = {arXiv},
  doi = {10.48550/arXiv.2211.10435},
  urldate = {2026-05-25},
  archiveprefix = {arXiv}
}

@inproceedings{gouCRITICLargeLanguage2023,
  title = {{{CRITIC}}: {{Large Language Models Can Self-Correct}} with {{Tool-Interactive Critiquing}}},
  shorttitle = {{{CRITIC}}},
  booktitle = {The {{Twelfth International Conference}} on {{Learning Representations}}},
  author = {Gou, Zhibin and Shao, Zhihong and Gong, Yeyun and Shen, Yelong and Yang, Yujiu and Duan, Nan and Chen, Weizhu},
  year = 2023,
  month = oct,
  urldate = {2026-05-27},
  langid = {english}
}

@misc{greshakeNotWhatYouve2023,
  title = {Not What You've Signed up for: {{Compromising Real-World LLM-Integrated Applications}} with {{Indirect Prompt Injection}}},
  shorttitle = {Not What You've Signed up For},
  author = {Greshake, Kai and Abdelnabi, Sahar and Mishra, Shailesh and Endres, Christoph and Holz, Thorsten and Fritz, Mario},
  year = 2023,
  month = may,
  number = {arXiv:2302.12173},
  eprint = {2302.12173},
  primaryclass = {cs.CR},
  publisher = {arXiv},
  doi = {10.48550/arXiv.2302.12173},
  urldate = {2026-05-25},
  archiveprefix = {arXiv}
}

@article{tsuiSelfCorrectionBench2025,
  title        = {{Self-Correction Bench: Uncovering and Addressing the Self-Correction Blind Spot in Large Language Models}},
  author       = {Tsui, Ken},
  year         = {2025},
  journal      = {arXiv preprint arXiv:2507.02778},
  eprint       = {2507.02778},
  archiveprefix = {arXiv},
  primaryclass = {cs.CL},
  url          = {https://arxiv.org/abs/2507.02778}
}

@inproceedings{kumarTrainingLanguageModels2024,
  title     = {{Training Language Models to Self-Correct via Reinforcement Learning}},
  author    = {Kumar, Aviral and Zhuang, Vincent and Agarwal, Rishabh and others},
  booktitle = {International Conference on Learning Representations (ICLR)},
  year      = {2025},
  eprint    = {2409.12917},
  archiveprefix = {arXiv},
  primaryclass  = {cs.LG}
}

@article{deepseekaiDeepSeekR12025,
  title   = {{DeepSeek-R1: Incentivizing Reasoning Capability in LLMs via Reinforcement Learning}},
  author  = {{DeepSeek-AI}},
  year    = {2025},
  journal = {arXiv preprint arXiv:2501.12948},
  eprint  = {2501.12948}, archiveprefix = {arXiv}, primaryclass = {cs.CL}
}

@article{maS2RTeaching2025,
  title   = {{S$^2$R: Teaching LLMs to Self-verify and Self-correct via Reinforcement Learning}},
  author  = {Ma, Ruotian and others},
  year    = {2025},
  journal = {arXiv preprint arXiv:2502.12853},
  eprint  = {2502.12853}, archiveprefix = {arXiv}, primaryclass = {cs.CL}
}

@misc{huangLargeLanguageModels2024,
  title = {Large {{Language Models Cannot Self-Correct Reasoning Yet}}},
  author = {Huang, Jie and Chen, Xinyun and Mishra, Swaroop and Zheng, Huaixiu Steven and Yu, Adams Wei and Song, Xinying and Zhou, Denny},
  year = 2024,
  month = mar,
  number = {arXiv:2310.01798},
  eprint = {2310.01798},
  primaryclass = {cs.CL},
  publisher = {arXiv},
  doi = {10.48550/arXiv.2310.01798},
  urldate = {2026-05-25},
  archiveprefix = {arXiv}
}

@misc{jiSurveyHallucinationNatural2022,
  title = {Survey of {{Hallucination}} in {{Natural Language Generation}}},
  author = {Ji, Ziwei and Lee, Nayeon and Frieske, Rita and Yu, Tiezheng and Su, Dan and Xu, Yan and Ishii, Etsuko and Bang, Yejin and Chen, Delong and Dai, Wenliang and Chan, Ho Shu and Madotto, Andrea and Fung, Pascale},
  year = 2022,
  month = feb,
  journal = {arXiv.org},
  doi = {10.1145/3571730},
  urldate = {2026-05-25},
  howpublished = {https://arxiv.org/abs/2202.03629v7},
  langid = {english}
}

@misc{kamoiWhenCanLLMs2024b,
  title = {When {{Can LLMs Actually Correct Their Own Mistakes}}? {{A Critical Survey}} of {{Self-Correction}} of {{LLMs}}},
  shorttitle = {When {{Can LLMs Actually Correct Their Own Mistakes}}?},
  author = {Kamoi, Ryo and Zhang, Yusen and Zhang, Nan and Han, Jiawei and Zhang, Rui},
  year = 2024,
  month = dec,
  eprint = {2406.01297},
  primaryclass = {cs.CL},
  doi = {10.1162/tacl_a_00713/125177},
  urldate = {2026-05-25},
  archiveprefix = {arXiv}
}

@misc{kimChallengingEvaluatorLLM2025,
  title = {Challenging the {{Evaluator}}: {{LLM Sycophancy Under User Rebuttal}}},
  shorttitle = {Challenging the {{Evaluator}}},
  author = {Kim, Sungwon and Khashabi, Daniel},
  year = 2025,
  month = sep,
  number = {arXiv:2509.16533},
  eprint = {2509.16533},
  primaryclass = {cs.CL},
  publisher = {arXiv},
  doi = {10.48550/arXiv.2509.16533},
  urldate = {2026-05-25},
  archiveprefix = {arXiv}
}

@misc{kojimaLargeLanguageModels2023,
  title = {Large {{Language Models}} Are {{Zero-Shot Reasoners}}},
  author = {Kojima, Takeshi and Gu, Shixiang Shane and Reid, Machel and Matsuo, Yutaka and Iwasawa, Yusuke},
  year = 2023,
  month = jan,
  number = {arXiv:2205.11916},
  eprint = {2205.11916},
  primaryclass = {cs.CL},
  publisher = {arXiv},
  doi = {10.48550/arXiv.2205.11916},
  urldate = {2026-05-25},
  archiveprefix = {arXiv}
}

@misc{lanhamMeasuringFaithfulnessChainofThought2023,
  title = {Measuring {{Faithfulness}} in {{Chain-of-Thought Reasoning}}},
  author = {Lanham, Tamera and Chen, Anna and Radhakrishnan, Ansh and Steiner, Benoit and Denison, Carson and Hernandez, Danny and Li, Dustin and Durmus, Esin and Hubinger, Evan and Kernion, Jackson and Luko{\v s}i{\=u}t{\.e}, Kamil{\.e} and Nguyen, Karina and Cheng, Newton and Joseph, Nicholas and Schiefer, Nicholas and Rausch, Oliver and Larson, Robin and McCandlish, Sam and Kundu, Sandipan and Kadavath, Saurav and Yang, Shannon and Henighan, Thomas and Maxwell, Timothy and {Telleen-Lawton}, Timothy and Hume, Tristan and {Hatfield-Dodds}, Zac and Kaplan, Jared and Brauner, Jan and Bowman, Samuel R. and Perez, Ethan},
  year = 2023,
  month = jul,
  number = {arXiv:2307.13702},
  eprint = {2307.13702},
  primaryclass = {cs.AI},
  publisher = {arXiv},
  doi = {10.48550/arXiv.2307.13702},
  urldate = {2026-05-25},
  archiveprefix = {arXiv}
}

@inproceedings{linZebraLogicScalingLimits2025,
  title = {{{ZebraLogic}}: {{On}} the {{Scaling Limits}} of {{LLMs}} for {{Logical Reasoning}}},
  shorttitle = {{{ZebraLogic}}},
  booktitle = {Forty-Second {{International Conference}} on {{Machine Learning}}},
  author = {Lin, Bill Yuchen and Bras, Ronan Le and Richardson, Kyle and Sabharwal, Ashish and Poovendran, Radha and Clark, Peter and Choi, Yejin},
  year = 2025,
  month = jun,
  urldate = {2026-05-27},
  langid = {english}
}

@misc{madaanSelfRefineIterativeRefinement2023,
  title = {Self-{{Refine}}: {{Iterative Refinement}} with {{Self-Feedback}}},
  shorttitle = {Self-{{Refine}}},
  author = {Madaan, Aman and Tandon, Niket and Gupta, Prakhar and Hallinan, Skyler and Gao, Luyu and Wiegreffe, Sarah and Alon, Uri and Dziri, Nouha and Prabhumoye, Shrimai and Yang, Yiming and Gupta, Shashank and Majumder, Bodhisattwa Prasad and Hermann, Katherine and Welleck, Sean and Yazdanbakhsh, Amir and Clark, Peter},
  year = 2023,
  month = may,
  number = {arXiv:2303.17651},
  eprint = {2303.17651},
  primaryclass = {cs.CL},
  publisher = {arXiv},
  doi = {10.48550/arXiv.2303.17651},
  urldate = {2026-05-25},
  archiveprefix = {arXiv}
}

@misc{olaussonSelfRepairSilverBullet2024,
  title = {Is {{Self-Repair}} a {{Silver Bullet}} for {{Code Generation}}?},
  author = {Olausson, Theo X. and Inala, Jeevana Priya and Wang, Chenglong and Gao, Jianfeng and {Solar-Lezama}, Armando},
  year = 2024,
  month = feb,
  number = {arXiv:2306.09896},
  eprint = {2306.09896},
  primaryclass = {cs.CL},
  publisher = {arXiv},
  doi = {10.48550/arXiv.2306.09896},
  urldate = {2026-05-25},
  archiveprefix = {arXiv}
}

@misc{openaiGPT4oSystemCard2024,
  title = {{{GPT-4o System Card}}},
  author = {{OpenAI}},
  year = 2024,
  month = oct,
  journal = {arXiv.org},
  urldate = {2026-05-25},
  howpublished = {https://arxiv.org/abs/2410.21276v1},
  langid = {english}
}

@misc{openaiGptOssModelCard2025,
  title = {{{gpt-oss-120b}} \& {{gpt-oss-20b Model Card}}},
  author = {{OpenAI}},
  year = 2025,
  month = aug,
  number = {arXiv:2508.10925},
  eprint = {2508.10925},
  primaryclass = {cs.CL},
  publisher = {arXiv},
  doi = {10.48550/arXiv.2508.10925},
  urldate = {2026-05-25},
  archiveprefix = {arXiv}
}

@misc{panUserAssistantBiasLLMs2026,
  title = {User-{{Assistant Bias}} in {{LLMs}}},
  author = {Pan, Xu and Fan, Jingxuan and Xiong, Zidi and Hahami, Ely and Overwiening, Jorin and Xie, Ziqian},
  year = 2026,
  month = apr,
  number = {arXiv:2508.15815},
  eprint = {2508.15815},
  primaryclass = {cs.CL},
  publisher = {arXiv},
  doi = {10.48550/arXiv.2508.15815},
  urldate = {2026-05-25},
  archiveprefix = {arXiv}
}

@misc{parkGenerativeAgentsInteractive2023,
  title = {Generative {{Agents}}: {{Interactive Simulacra}} of {{Human Behavior}}},
  shorttitle = {Generative {{Agents}}},
  author = {Park, Joon Sung and O'Brien, Joseph C. and Cai, Carrie J. and Morris, Meredith Ringel and Liang, Percy and Bernstein, Michael S.},
  year = 2023,
  month = aug,
  number = {arXiv:2304.03442},
  eprint = {2304.03442},
  primaryclass = {cs.HC},
  publisher = {arXiv},
  doi = {10.48550/arXiv.2304.03442},
  urldate = {2026-05-25},
  archiveprefix = {arXiv}
}

@misc{perezDiscoveringLanguageModel2022,
  title = {Discovering {{Language Model Behaviors}} with {{Model-Written Evaluations}}},
  author = {Perez, Ethan and Ringer, Sam and Luko{\v s}i{\=u}t{\.e}, Kamil{\.e} and Nguyen, Karina and Chen, Edwin and Heiner, Scott and Pettit, Craig and Olsson, Catherine and Kundu, Sandipan and Kadavath, Saurav and Jones, Andy and Chen, Anna and Mann, Ben and Israel, Brian and Seethor, Bryan and McKinnon, Cameron and Olah, Christopher and Yan, Da and Amodei, Daniela and Amodei, Dario and Drain, Dawn and Li, Dustin and {Tran-Johnson}, Eli and Khundadze, Guro and Kernion, Jackson and Landis, James and Kerr, Jamie and Mueller, Jared and Hyun, Jeeyoon and Landau, Joshua and Ndousse, Kamal and Goldberg, Landon and Lovitt, Liane and Lucas, Martin and Sellitto, Michael and Zhang, Miranda and Kingsland, Neerav and Elhage, Nelson and Joseph, Nicholas and Mercado, Noem{\'i} and DasSarma, Nova and Rausch, Oliver and Larson, Robin and McCandlish, Sam and Johnston, Scott and Kravec, Shauna and Showk, Sheer El and Lanham, Tamera and {Telleen-Lawton}, Timothy and Brown, Tom and Henighan, Tom and Hume, Tristan and Bai, Yuntao and {Hatfield-Dodds}, Zac and Clark, Jack and Bowman, Samuel R. and Askell, Amanda and Grosse, Roger and Hernandez, Danny and Ganguli, Deep and Hubinger, Evan and Schiefer, Nicholas and Kaplan, Jared},
  year = 2022,
  month = dec,
  number = {arXiv:2212.09251},
  eprint = {2212.09251},
  primaryclass = {cs.CL},
  publisher = {arXiv},
  doi = {10.48550/arXiv.2212.09251},
  urldate = {2026-05-25},
  archiveprefix = {arXiv}
}

@misc{qwenQwen25TechnicalReport2025,
  title = {Qwen2.5 {{Technical Report}}},
  author = {{Qwen}},
  year = 2025,
  month = jan,
  number = {arXiv:2412.15115},
  eprint = {2412.15115},
  primaryclass = {cs.CL},
  publisher = {arXiv},
  doi = {10.48550/arXiv.2412.15115},
  urldate = {2026-05-25},
  archiveprefix = {arXiv}
}

@misc{dhuliawalaChainofVerificationReducesHallucination2023,
  title = {Chain-of-Verification Reduces Hallucination in Large Language Models},
  author = {Dhuliawala, Shehzaad and Komeili, Mojtaba and Xu, Jing and Raileanu, Roberta and Li, Xian and Celikyilmaz, Asli and Weston, Jason},
  year = 2023,
  month = sep,
  number = {arXiv:2309.11495},
  eprint = {2309.11495},
  archiveprefix = {arXiv},
  primaryclass = {cs.CL},
  publisher = {arXiv},
  doi = {10.48550/arXiv.2309.11495}
}

@misc{qwenQwen3TechnicalReport2025,
  title = {Qwen3 {{Technical Report}}},
  author = {{Qwen}},
  year = 2025,
  month = may,
  number = {arXiv:2505.09388},
  eprint = {2505.09388},
  primaryclass = {cs.CL},
  publisher = {arXiv},
  doi = {10.48550/arXiv.2505.09388},
  urldate = {2026-05-25},
  archiveprefix = {arXiv}
}

@misc{schickToolformerLanguageModels2023,
  title = {Toolformer: {{Language Models Can Teach Themselves}} to {{Use Tools}}},
  shorttitle = {Toolformer},
  author = {Schick, Timo and {Dwivedi-Yu}, Jane and Dess{\`i}, Roberto and Raileanu, Roberta and Lomeli, Maria and Zettlemoyer, Luke and Cancedda, Nicola and Scialom, Thomas},
  year = 2023,
  month = feb,
  number = {arXiv:2302.04761},
  eprint = {2302.04761},
  primaryclass = {cs.CL},
  publisher = {arXiv},
  doi = {10.48550/arXiv.2302.04761},
  urldate = {2026-05-25},
  archiveprefix = {arXiv}
}

@misc{sharmaUnderstandingSycophancyLanguage2025,
  title = {Towards {{Understanding Sycophancy}} in {{Language Models}}},
  author = {Sharma, Mrinank and Tong, Meg and Korbak, Tomasz and Duvenaud, David and Askell, Amanda and Bowman, Samuel R. and Cheng, Newton and Durmus, Esin and {Hatfield-Dodds}, Zac and Johnston, Scott R. and Kravec, Shauna and Maxwell, Timothy and McCandlish, Sam and Ndousse, Kamal and Rausch, Oliver and Schiefer, Nicholas and Yan, Da and Zhang, Miranda and Perez, Ethan},
  year = 2025,
  month = may,
  number = {arXiv:2310.13548},
  eprint = {2310.13548},
  primaryclass = {cs.CL},
  publisher = {arXiv},
  doi = {10.48550/arXiv.2310.13548},
  urldate = {2026-05-25},
  archiveprefix = {arXiv}
}

@inproceedings{shinnReflexionLanguageAgents2023,
  title = {Reflexion: Language Agents with Verbal Reinforcement Learning},
  shorttitle = {Reflexion},
  booktitle = {Thirty-Seventh {{Conference}} on {{Neural Information Processing Systems}}},
  author = {Shinn, Noah and Cassano, Federico and Gopinath, Ashwin and Narasimhan, Karthik R. and Yao, Shunyu},
  year = 2023,
  month = nov,
  urldate = {2026-05-27},
  langid = {english}
}

@misc{srivastavaImitationGameQuantifying2023,
  title = {Beyond the Imitation Game: Quantifying and Extrapolating the Capabilities of Language Models},
  shorttitle = {Beyond the Imitation Game},
  author = {{BIG-bench Collaboration}},
  year = 2023,
  month = jun,
  number = {arXiv:2206.04615},
  eprint = {2206.04615},
  archiveprefix = {arXiv},
  primaryclass = {cs.CL},
  publisher = {arXiv},
  doi = {10.48550/arXiv.2206.04615}
}

@misc{stechlyGPT4DoesntKnow2023a,
  title = {{{GPT-4 Doesn}}'t {{Know It}}'s {{Wrong}}: {{An Analysis}} of {{Iterative Prompting}} for {{Reasoning Problems}}},
  shorttitle = {{{GPT-4 Doesn}}'t {{Know It}}'s {{Wrong}}},
  author = {Stechly, Kaya and Marquez, Matthew and Kambhampati, Subbarao},
  year = 2023,
  month = oct,
  number = {arXiv:2310.12397},
  eprint = {2310.12397},
  primaryclass = {cs.AI},
  publisher = {arXiv},
  doi = {10.48550/arXiv.2310.12397},
  urldate = {2026-05-25},
  archiveprefix = {arXiv}
}

@misc{suzgunChallengingBIGBenchTasks2022,
  title = {Challenging {{BIG-Bench Tasks}} and {{Whether Chain-of-Thought Can Solve Them}}},
  author = {Suzgun, Mirac and Scales, Nathan and Sch{\"a}rli, Nathanael and Gehrmann, Sebastian and Tay, Yi and Chung, Hyung Won and Chowdhery, Aakanksha and Le, Quoc V. and Chi, Ed H. and Zhou, Denny and Wei, Jason},
  year = 2022,
  month = oct,
  number = {arXiv:2210.09261},
  eprint = {2210.09261},
  primaryclass = {cs.CL},
  publisher = {arXiv},
  doi = {10.48550/arXiv.2210.09261},
  urldate = {2026-05-25},
  archiveprefix = {arXiv}
}

@misc{teamGemma3Technical2025,
  title = {Gemma 3 {{Technical Report}}},
  author = {{Gemma Team}},
  year = 2025,
  month = mar,
  number = {arXiv:2503.19786},
  eprint = {2503.19786},
  primaryclass = {cs.CL},
  publisher = {arXiv},
  doi = {10.48550/arXiv.2503.19786},
  urldate = {2026-05-25},
  archiveprefix = {arXiv}
}

@misc{turpinLanguageModelsDont2023,
  title = {Language {{Models Don}}'t {{Always Say What They Think}}: {{Unfaithful Explanations}} in {{Chain-of-Thought Prompting}}},
  shorttitle = {Language {{Models Don}}'t {{Always Say What They Think}}},
  author = {Turpin, Miles and Michael, Julian and Perez, Ethan and Bowman, Samuel R.},
  year = 2023,
  month = dec,
  number = {arXiv:2305.04388},
  eprint = {2305.04388},
  primaryclass = {cs.CL},
  publisher = {arXiv},
  doi = {10.48550/arXiv.2305.04388},
  urldate = {2026-05-25},
  archiveprefix = {arXiv}
}

@misc{tyenLLMsCannotFind2024,
  title = {{{LLMs}} Cannot Find Reasoning Errors, but Can Correct Them given the Error Location},
  author = {Tyen, Gladys and Mansoor, Hassan and C{\u a}rbune, Victor and Chen, Peter and Mak, Tony},
  year = 2024,
  month = jun,
  number = {arXiv:2311.08516},
  eprint = {2311.08516},
  primaryclass = {cs.AI},
  publisher = {arXiv},
  doi = {10.48550/arXiv.2311.08516},
  urldate = {2026-05-25},
  archiveprefix = {arXiv}
}

@misc{wallaceInstructionHierarchyTraining2024,
  title = {The {{Instruction Hierarchy}}: {{Training LLMs}} to {{Prioritize Privileged Instructions}}},
  shorttitle = {The {{Instruction Hierarchy}}},
  author = {Wallace, Eric and Xiao, Kai and Leike, Reimar and Weng, Lilian and Heidecke, Johannes and Beutel, Alex},
  year = 2024,
  month = apr,
  number = {arXiv:2404.13208},
  eprint = {2404.13208},
  primaryclass = {cs.CR},
  publisher = {arXiv},
  doi = {10.48550/arXiv.2404.13208},
  urldate = {2026-05-25},
  archiveprefix = {arXiv}
}

@misc{wangSelfConsistencyImprovesChain2023,
  title = {Self-{{Consistency Improves Chain}} of {{Thought Reasoning}} in {{Language Models}}},
  author = {Wang, Xuezhi and Wei, Jason and Schuurmans, Dale and Le, Quoc and Chi, Ed and Narang, Sharan and Chowdhery, Aakanksha and Zhou, Denny},
  year = 2023,
  month = mar,
  number = {arXiv:2203.11171},
  eprint = {2203.11171},
  primaryclass = {cs.CL},
  publisher = {arXiv},
  doi = {10.48550/arXiv.2203.11171},
  urldate = {2026-05-25},
  archiveprefix = {arXiv}
}

@misc{weiAMemGuardProactiveDefense2025,
  title = {A-{{MemGuard}}: {{A Proactive Defense Framework}} for {{LLM-Based Agent Memory}}},
  shorttitle = {A-{{MemGuard}}},
  author = {Wei, Qianshan and Yang, Tengchao and Wang, Yaochen and Li, Xinfeng and Li, Lijun and Yin, Zhenfei and Zhan, Yi and Holz, Thorsten and Lin, Zhiqiang and Wang, XiaoFeng},
  year = 2025,
  month = sep,
  number = {arXiv:2510.02373},
  eprint = {2510.02373},
  primaryclass = {cs.CR},
  publisher = {arXiv},
  doi = {10.48550/arXiv.2510.02373},
  urldate = {2026-05-25},
  archiveprefix = {arXiv}
}

@misc{weiChainofThoughtPromptingElicits2023,
  title = {Chain-of-{{Thought Prompting Elicits Reasoning}} in {{Large Language Models}}},
  author = {Wei, Jason and Wang, Xuezhi and Schuurmans, Dale and Bosma, Maarten and Ichter, Brian and Xia, Fei and Chi, Ed and Le, Quoc and Zhou, Denny},
  year = 2023,
  month = jan,
  number = {arXiv:2201.11903},
  eprint = {2201.11903},
  primaryclass = {cs.CL},
  publisher = {arXiv},
  doi = {10.48550/arXiv.2201.11903},
  urldate = {2026-05-25},
  archiveprefix = {arXiv}
}

@misc{weiSimpleSyntheticData2024,
  title = {Simple Synthetic Data Reduces Sycophancy in Large Language Models},
  author = {Wei, Jerry and Huang, Da and Lu, Yifeng and Zhou, Denny and Le, Quoc V.},
  year = 2024,
  month = feb,
  number = {arXiv:2308.03958},
  eprint = {2308.03958},
  primaryclass = {cs.CL},
  publisher = {arXiv},
  doi = {10.48550/arXiv.2308.03958},
  urldate = {2026-05-25},
  archiveprefix = {arXiv}
}

@inproceedings{welleckGeneratingSequencesLearning2022,
  title = {{Generating Sequences by Learning to Self-Correct}},
  booktitle = {{The Eleventh International Conference on Learning Representations}},
  author = {Welleck, Sean and Lu, Ximing and West, Peter and Brahman, Faeze and Shen, Tianxiao and Khashabi, Daniel and Choi, Yejin},
  year = 2022,
  month = sep,
  urldate = {2026-05-27},
  langid = {chinese-traditional}
}

@misc{yaoReActSynergizingReasoning2023,
  title = {{{ReAct}}: {{Synergizing Reasoning}} and {{Acting}} in {{Language Models}}},
  shorttitle = {{{ReAct}}},
  author = {Yao, Shunyu and Zhao, Jeffrey and Yu, Dian and Du, Nan and Shafran, Izhak and Narasimhan, Karthik and Cao, Yuan},
  year = 2023,
  month = mar,
  number = {arXiv:2210.03629},
  eprint = {2210.03629},
  primaryclass = {cs.CL},
  publisher = {arXiv},
  doi = {10.48550/arXiv.2210.03629},
  urldate = {2026-05-25},
  archiveprefix = {arXiv}
}

@misc{yaoTreeThoughtsDeliberate2023,
  title = {Tree of {{Thoughts}}: {{Deliberate Problem Solving}} with {{Large Language Models}}},
  shorttitle = {Tree of {{Thoughts}}},
  author = {Yao, Shunyu and Yu, Dian and Zhao, Jeffrey and Shafran, Izhak and Griffiths, Thomas L. and Cao, Yuan and Narasimhan, Karthik},
  year = 2023,
  month = dec,
  number = {arXiv:2305.10601},
  eprint = {2305.10601},
  primaryclass = {cs.CL},
  publisher = {arXiv},
  doi = {10.48550/arXiv.2305.10601},
  urldate = {2026-05-27},
  archiveprefix = {arXiv}
}

@misc{yinReasoningTrapHow2026,
  title = {The {{Reasoning Trap}}: {{How Enhancing LLM Reasoning Amplifies Tool Hallucination}}},
  shorttitle = {The {{Reasoning Trap}}},
  author = {Yin, Chenlong and Sha, Zeyang and Cui, Shiwen and Meng, Changhua and Li, Zechao},
  year = 2026,
  month = apr,
  number = {arXiv:2510.22977},
  eprint = {2510.22977},
  primaryclass = {cs.LG},
  publisher = {arXiv},
  doi = {10.48550/arXiv.2510.22977},
  urldate = {2026-05-25},
  archiveprefix = {arXiv}
}

@article{zhangSirensSongAI2025,
  title = {{{Siren}}'s {{Song}} in the {{AI Ocean}}: {{A Survey}} on {{Hallucination}} in {{Large Language Models}}},
  author = {Zhang, Yue and Li, Yafu and Cui, Leyang and Cai, Deng and Liu, Lemao and Fu, Tingchen and Huang, Xinting and Zhao, Enbo and Zhang, Yu and Chen, Yulong and Wang, Longyue and Luu, Anh Tuan and Bi, Wei and Shi, Freda and Shi, Shuming},
  year = 2025,
  month = feb,
  journal = {Computational Linguistics},
  volume = {51},
  number = {4},
  pages = {1373--1418},
  publisher = {MIT Press},
  address = {Cambridge, MA},
  doi = {10.1162/coli.a.16},
  urldate = {2026-05-25}
}

@misc{zhangSurveyMemoryMechanism2024,
  title = {A {{Survey}} on the {{Memory Mechanism}} of {{Large Language Model}} Based {{Agents}}},
  author = {Zhang, Zeyu and Bo, Xiaohe and Ma, Chen and Li, Rui and Chen, Xu and Dai, Quanyu and Zhu, Jieming and Dong, Zhenhua and Wen, Ji-Rong},
  year = 2024,
  month = apr,
  number = {arXiv:2404.13501},
  eprint = {2404.13501},
  primaryclass = {cs.AI},
  publisher = {arXiv},
  doi = {10.48550/arXiv.2404.13501},
  urldate = {2026-05-25},
  archiveprefix = {arXiv}
}

@misc{zhaoBoostingLLMReasoning2025,
  title = {Boosting {{LLM Reasoning}} via {{Spontaneous Self-Correction}}},
  author = {Zhao, Xutong and Xu, Tengyu and Wang, Xuewei and Chen, Zhengxing and Jin, Di and Tan, Liang and {Yen-Ting} and Yu, Zishun and Zhao, Zhuokai and He, Yun and Wang, Sinong and Fang, Han and Chandar, Sarath and Zhu, Chen},
  year = 2025,
  month = jun,
  number = {arXiv:2506.06923},
  eprint = {2506.06923},
  primaryclass = {cs.AI},
  publisher = {arXiv},
  doi = {10.48550/arXiv.2506.06923},
  urldate = {2026-05-25},
  archiveprefix = {arXiv}
}

\appendix
\newpage
%%%%%%%%%%%%%%%%%%%%%%%%%%%%%%%%%%%%%%%%%%%%%%%%%%%%%%%%%%%%%%%%%%%%%%%
%%%%%%%%%%%%%%%%%%%%%%%%%% A: Statistical robustness %%%%%%%%%%%%%%%%%%
%%%%%%%%%%%%%%%%%%%%%%%%%%%%%%%%%%%%%%%%%%%%%%%%%%%%%%%%%%%%%%%%%%%%%%%
\section{Statistical Robustness}
\label{app:stats}

\subsection{Multiple-Comparison Correction}
\label{app:stats-multicomp}

The main table reports 48 paired contrasts (12 combinations $\times$ 4 relabel conditions). We apply Holm-Bonferroni at family-wise error rate $\alpha=0.05$. Of the 48 contrasts, 28 survive, and 9 of 12 combinations retain at least one significant contrast. The three combinations without a surviving significant contrast are the two with a saturated baseline (gpt-oss-20B math and Qwen-72B BBH-LD), which carry no significant contrast before correction either, plus Gemini 2.5 Flash-Lite, whose only significant contrast was the borderline \cond{L\_memory} entry at $\Delta = +23.3$\,pp. The headline finding (10 of 12 combinations significant before correction; 9 of 12 after Holm-Bonferroni) is therefore robust to strict multiple-testing correction.

\subsection{Failure-Pool Stochasticity}
\label{app:stats-stochasticity}

The \cond{L0\_self} baseline is measured at $T=0$ with a single seed, so we ask whether the $+53$\,pp increase on \cond{L\_memory} could reflect sampling variability rather than the role manipulation. On the same 30 Qwen-72B math failure-pool tasks, we re-run \cond{L0\_self} at $T=0.5$ with five independent seeds, holding the prior trajectory and \cstar{} fixed. The aggregate re-sampled correction rate is $16.0\%$ (95\% CI $[8.7, 24.7]$), indistinguishable from the main-run $16.7\%$ (per-seed $10.0$--$20.0\%$). The \cond{L\_memory} gain over this baseline is $+54.0$\,pp, slightly larger than over the single-seed one. The increase is not attributable to sampling variability.

\subsection{Multi-Judge Robustness}
\label{app:stats-multijudge}

The headline analyses use a locked Qwen-72B judge. This judge could in principle be biased toward Qwen-style explicit identification, so that responses from non-Qwen agents (Llama, GPT-4o, Claude, Gemini) are judged differently. To test this, we stratify a sample of 200 trials (40 per source) across five model classes (Qwen-72B math, Llama-3.3-70B math, GPT-4o math, Claude Sonnet 4 math, Gemini 2.5 Flash-Lite math) and re-judge each trajectory with a second judge (Llama-3.3-70B) using the same locked judge prompt.

Two judges agree on $92.5\%$ of 200 trials with Cohen's $\kappa = 0.843$. The confusion matrix is dominated by joint-YES ($115$) and joint-NO ($70$); disagreements are asymmetric, with Qwen-72B labeling YES while Llama-3.3-70B labels NO 14 times, and the reverse only 1 time. Qwen-72B is therefore the slightly more generous judge, so swapping in Llama-3.3-70B as the canonical judge would lower a few correction rates but preserve rank ordering. Agreement is lowest where Qwen judges Qwen ($82.5\%$), so the criterion is stable in ordering rather than judge-invariant.

\subsection{Final-Answer Correctness}
\label{app:final-answer-metric}

The strict-identification criterion used throughout the paper measures whether the agent explicitly names \cstar{} and rejects it. By construction this metric may under-credit silent re-derivation or, conversely, mask a final-answer regression. To probe both, we re-score every Qwen-72B math and Llama-3.3-70B math trial with the locked Qwen-72B judge under a different prompt: ``given the gold answer $X$, did the response state $X$ as its final answer, regardless of phrasing?''. We use a judge rather than a regex because reword\-ed phrasings such as ``the correct answer is 42'' or ``so it's 42'' miss a literal ``Final Answer: 42'' match yet still arrive at the correct numeric value, and the miss is not uniform across conditions.

\begin{table}[t]
\centering
\small
\caption{Multi-judge robustness on 200 stratified trials. Two judges (Qwen-72B and Llama-3.3-70B) re-judge the same trajectories with the same locked prompt; agreement is computed on the strict-identification verdict (YES/NO).}
\label{tab:appendix-multijudge}
\begin{tabular*}{\columnwidth}{@{\extracolsep{\fill}} l r r @{}}
\toprule
\textbf{Source} & $n$ & Agreement \\
\midrule
Qwen-72B main           & 40 & 82.5\% \\
Llama-3.3-70B main      & 40 & 97.5\% \\
GPT-4o main             & 40 & 95.0\% \\
Claude Sonnet 4 main    & 40 & 97.5\% \\
Gemini 2.5 Flash-Lite   & 40 & 90.0\% \\
\midrule
\textbf{Overall}        & \textbf{200} & \textbf{92.5\%} \\
\multicolumn{3}{@{}l@{}}{\textbf{Cohen's $\kappa = 0.843$ (almost perfect agreement)}} \\
\bottomrule
\end{tabular*}
\end{table}

\begin{table}[t]
\centering
\small
\setlength{\tabcolsep}{4pt}
\caption{Strict-identification CR vs.\ judge-verified final-answer correctness ($\%$) on the two 70B-class math failure pools ($n=30$ paired). $\Delta$ final is the paired contrast of final CR vs.\ \cond{L0\_self}; all $p>0.05$ (paired bootstrap, two-sided).}
\label{tab:appendix-final-answer}
\begin{tabular*}{\columnwidth}{@{\extracolsep{\fill}} l l r r r @{}}
\toprule
\textbf{Model} & \textbf{Cond.} & strict & final & $\Delta$ final \\
\midrule
\multirow{5}{*}{Qwen-72B}  & \cond{L0\_self}        & 16.7 & 76.7 & --- \\
                           & \cond{L\_user\_wait}   & 40.0 & 80.0 & $+3.3$ \\
                           & \cond{L\_user\_neutral}& 43.3 & 86.7 & $+10.0$ \\
                           & \cond{L\_tool}         & 43.3 & 76.7 & $\pm 0$ \\
                           & \cond{L\_memory}       & 70.0 & 80.0 & $+3.3$ \\
\midrule
\multirow{5}{*}{Llama-70B} & \cond{L0\_self}        & 0.0  & 70.0 & --- \\
                           & \cond{L\_user\_wait}   & 56.7 & 73.3 & $+3.3$ \\
                           & \cond{L\_user\_neutral}& 60.0 & 80.0 & $+10.0$ \\
                           & \cond{L\_tool}         & 3.3  & 86.7 & $+16.7$ \\
                           & \cond{L\_memory}       & 86.7 & 56.7 & $-13.3$ \\
\bottomrule
\end{tabular*}
\end{table}

Three observations follow. First, on both models the audit-only baseline already reaches the correct final answer in $70$ to $77\%$ of trials despite a strict CR of $0$ to $17\%$, evidencing strong silent re-derivation that the strict criterion explicitly does not credit. Second, across all eight relabel-vs-baseline contrasts the paired final-CR difference falls in $[-13, +17]$\,pp and none is significant at $p<0.05$; the strict-CR increases do not come at the cost of final-answer accuracy. Third, the two metrics rank conditions differently: on Llama, \cond{L\_tool} has the highest final CR ($86.7\%$) but a near-zero strict CR ($3.3\%$), and \cond{L\_memory} reverses this with the highest strict CR ($86.7\%$) and the lowest final CR ($56.7\%$, paired $\Delta=-13.3$\,pp, CI $[-33, +7]$). The non-significant negative trend on \cond{L\_memory} is consistent with a known failure mode: an agent that explicitly commits to a replacement value for \cstar{} occasionally locks in an incorrect candidate, whereas silent re-derivation is free to re-roll. The two metrics are therefore complementary; the main paper reports strict CR because addressability is the question this paper asks, and the absence of a significant final-CR regression bounds the deployment risk.

%%%%%%%%%%%%%%%%%%%%%%%%%%%%%%%%%%%%%%%%%%%%%%%%%%%%%%%%%%%%%%%%%%%%%%%
%%%%%%%%%%%%%%%%% B: Comparison with published methods %%%%%%%%%%%%%%%%
%%%%%%%%%%%%%%%%%%%%%%%%%%%%%%%%%%%%%%%%%%%%%%%%%%%%%%%%%%%%%%%%%%%%%%%
\section{Comparison with Published Self-Correction Methods}
\label{app:baselines}

We test whether the increase observed in the main table replicates when published intrinsic self-correction protocols are run on the same failure pool under the same locked judge. Three methods are compared against the audit-only baseline and the role-relabel reference: Self-Refine~\citep{madaanSelfRefineIterativeRefinement2023}, Reflexion~\citep{shinnReflexionLanguageAgents2023}, and Chain-of-Verification (CoVe)~\citep{dhuliawalaChainofVerificationReducesHallucination2023}.

\paragraph{Protocols.} The Self-Refine critique instruction is taken verbatim from the Madaan et al.\ GSM8K release, adapted by substituting ``reasoning'' for ``code'' since our solutions are natural-language traces rather than PaL-style code. Reflexion is run as a two-cycle reflect-and-retry following the HotpotQA pattern. CoVe is implemented as the three-step plan/execute/audit pipeline: the agent first plans verification questions about its intermediate values, each question is answered independently with a fresh-context call, and the verification answers are surfaced before the trailing audit. All four methods (the three baselines plus \cond{L\_memory}) follow the same audit prompt; only the wrapping context differs.

\begin{table}[t]
\centering
\small
\setlength{\tabcolsep}{4pt}
\caption{Comparison with published self-correction methods on the math failure pools of four models, under the same locked judge. Entries are correction rate (\%) with significance against each row's own audit-only baseline. Rate limits cap $n$ on the two closed-weight models, whose in-text increases are paired contrasts over each method's common-completion subset and need not equal these marginal-rate differences.}
\label{tab:appendix-baselines}
\begin{tabular*}{\columnwidth}{@{\extracolsep{\fill}} l r r r r r r @{}}
\toprule
Model & $n$ & audit & S.-Refine & Reflex. & CoVe & \cond{L\_mem.} \\
\midrule
\multicolumn{7}{@{}l}{\textit{Open-weight}} \\
\quad Qwen-72B  & 30 & 30.0 & 6.7 & 46.7 & 10.0 & $\mathbf{80.0^{***}}$ \\
\quad Llama-70B & 30 & 3.3 & 0.0 & 6.7 & 0.0 & $\mathbf{86.7^{***}}$ \\
\midrule
\multicolumn{7}{@{}l}{\textit{Closed-weight}} \\
\quad GPT-4o  & $\approx$18 & 31.6 & $63.2^{**}$ & $50.0^{*}$ & 11.1 & $\mathbf{72.2^{***}}$ \\
\quad GPT-4.1 & $\approx$17 & 82.4 & 76.5 & $\mathbf{100.0}$ & 35.3 & 93.8 \\
\bottomrule
\end{tabular*}
\end{table}

\paragraph{Open-weight models.} Table~\ref{tab:appendix-baselines} reports the four-method comparison on all four models. On the two open-weight math failure pools, Self-Refine and CoVe produce \emph{negative} changes under the strict-identification criterion: both methods induce generic re-derivation that arrives at correct final answers but never names the wrong intermediate, so the locked judge returns NO. Reflexion produces a small, non-significant positive change on both models, remaining near the audit-only baseline on Llama. The role-relabel reference dominates on both models. To preserve strict paired comparison, the audit-only condition and \cond{L\_memory} were re-run within this batch rather than imported. Batch-level non-determinism in the Ollama runtime at $T=0$ drifts their absolute values from the main table: on Qwen-72B the canonical $16.7\%$ and $70.0\%$ become $30.0\%$ and $80.0\%$. The within-batch paired contrast is still $+50.0$\,pp, agreeing with the canonical $+53.3$\,pp within bootstrap CI, and the method ordering and \cond{L\_memory} dominance hold across both batches.

\paragraph{Closed-weight models.} We replicate the comparison on two standard instruction-tuned closed-weight models, served via the GitHub Models API. Per-condition $n$ is smaller than $30$ because free-tier rate limits cap how many trials can be run within the experimental window. On GPT-4o the ordering replicates the role-relabel as the strongest single intervention ($+52.9$\,pp, $p<0.001$), with Self-Refine and Reflexion now also significant ($+31.6$ and $+22.2$\,pp) and CoVe still hurting. On GPT-4.1 the audit-only baseline already sits at $82.4\%$, a saturated baseline, so all methods cluster near the ceiling except CoVe, which drops to $35.3\%$ ($-47.1$\,pp) by fragmenting the response into verification questions the strict judge does not credit.

\paragraph{Take-away.} Under the strict-identification criterion, the role-relabel intervention dominates the published baselines on every model whose baseline is not saturated. The pattern is consistent with the addressability account: methods that drive the agent through additional reasoning passes solve the underlying arithmetic but fail to address the wrong intermediate as a discrete entity.
%%%%%%%%%%%%%%%%%%%%%%%%%%%%%%%%%%%%%%%%%%%%%%%%%%%%%%%%%%%%%%%%%%%%%%%
%%%%%%%%%%%%%%%%%%%%%% C: Mechanism triangulation %%%%%%%%%%%%%%%%%%%%%
%%%%%%%%%%%%%%%%%%%%%%%%%%%%%%%%%%%%%%%%%%%%%%%%%%%%%%%%%%%%%%%%%%%%%%%
\section{Mechanism Triangulation}
\label{app:mechanism}

\subsection{Role-Token Variants}
\label{app:mechanism-tokens}

The wrapper-role ladder of Sec.~\ref{sec:handle-granularity} isolates the role label from a bare syntactic wrapper but uses the canonical \role{<memory>} tag throughout. To separate the contribution of the system-role token pair from the memory-aligned semantic prior on the tag, we vary the tag while holding the role (\role{system}) and the wrapped content (\cstar) constant.

\begin{table}[t]
\centering
\small
\caption{Correction rate under different system-role tag wrappers on Qwen-72B math ($n=30$ paired). The right column is the paired contrast against \role{<memory>}, in pp; $^{***}p<0.001$ (two-sided paired bootstrap), unmarked entries are not significant (all $p>0.05$), as in Table~\ref{tab:final-main}.}
\label{tab:appendix-tokens}
\begin{tabular*}{\columnwidth}{@{\extracolsep{\fill}} l r r @{}}
\toprule
\textbf{Tag} & CR & $\Delta$ to \role{<memory>} \\
\midrule
\role{<memory>}             & \textbf{70.0\%} & --- \\
\role{<retrieved>}          & 63.3\%          & $-6.7\phantom{^{***}}$ \\
\role{<context>}            & 53.3\%          & $-16.7\phantom{^{***}}$ \\
\role{<note>}               & 50.0\%          & $-20.0\phantom{^{***}}$ \\
\role{<reference>}          & 53.3\%          & $-16.7\phantom{^{***}}$ \\
\role{<xqzy>} (nonsense)    & 30.0\%          & $-40.0^{***}$ \\
\midrule
(reference) \cond{L0\_self} & 16.7\%          & --- \\
\bottomrule
\end{tabular*}
\end{table}

The pattern decomposes into two layers. A system-role wrapper whose tag is semantically empty is worth no more than a bare syntactic boundary: the gibberish \role{<xqzy>} reaches $30.0\%$, $+13.3$\,pp above the \cond{L0\_self} baseline, but no higher than the label-free \cond{H1}--\cond{H3} rungs ($33.3$ to $40.0\%$, Table~\ref{tab:handle}). The tag's lexical identity carries the decisive share: \role{<memory>} dominates the gibberish control by $40.0$\,pp ($p=0.0002$, paired bootstrap), and is statistically indistinguishable from its closest neighbor \role{<retrieved>}. With the H0--H4 ladder, the picture is: a syntactic boundary supplies a baseline increase, and a memory-aligned semantic prior carried by a \role{system} role pushes it to the full $+53.3$\,pp.

\subsection{First-Token Signature Across Model Families}
\label{app:mechanism-tokens-models}

The engage-then-verify first-token signature reported in Sec.~\ref{sec:exp-addressability} is on Qwen-72B. To check whether the specific lexical mode generalizes, we replicate the logprob experiment on three additional standard instruction-tuned families: Llama-3.3-70B, Phi-4-14B, and Gemma-3-12B. Reasoning-mode models are excluded by construction, since a trained reasoning preamble (e.g.\ Qwen3-30B's ``Okay'') deterministically occupies the first-token slot and leaves no stance signal to read. Table~\ref{tab:appendix-logprobs-cross} reports the agreement-token mass under each role context. The lexical first-token mode is markedly model-specific: the \cond{L\_memory} agreement spike defining the Qwen-72B signature does not reproduce elsewhere, with Gemma-3 concentrating agreement mass under \cond{L\_user\_neutral}, Phi-4 under \cond{L0\_self}, and Llama-3.3-70B keeping agreement mass uniformly low. The engage-then-verify signature is thus, in its specific lexical form, a Qwen-72B feature; the behavioral increase (Table~\ref{tab:final-main}) holds for all four families, but the first-token route to rejection differs by tokenizer and training mix.

\begin{table}[t]
\centering
\small
\setlength{\tabcolsep}{3pt}
\caption{First-token mean probability mass on agreement tokens (Indeed, Yes, So, \ldots) by role context, across four open-weight families on the same 15-task failure-pool subset. Columns are the conditions \cond{L0\_self} (\cstar{} in \role{<thought>}), \cond{L\_user\_neutral}, \cond{L\_tool}, and \cond{L\_memory}. Gemma-3 has no native \role{tool} role, so its \cond{L\_tool} entry is not applicable (Appendix~\ref{app:chat-templates}).}
\label{tab:appendix-logprobs-cross}
\begin{tabular*}{\columnwidth}{@{\extracolsep{\fill}}lrrrr@{}}
\toprule
\textbf{Model} & \cond{L0\_self} & \cond{user} & \cond{tool} & \cond{mem.} \\
\midrule
Qwen-72B (ref.)  & 0.001          & 0.0001 & 0.0005         & \textbf{0.226} \\
Llama-3.3-70B         & 0.019          & 0.0001 & 0.017          & 0.004 \\
Phi-4-14B             & \textbf{0.212} & 0.0004 & 0.019          & 0.019 \\
Gemma-3-12B           & 0.005          & 0.067  & n/a & $\sim 0$ \\
\bottomrule
\end{tabular*}
\end{table}

\subsection{Hidden-State Probing}
\label{app:mechanism-probe}

We test whether the relabel produces a representation change that is linearly readable from the model's final-layer embedding. For each of the 30 Qwen-72B math failure-pool tasks and each of the five conditions, we render the trailing message slice (the role-wrapped \cstar{} plus the audit prompt) and embed it via the served model's embedding endpoint, yielding a $30 \times 5$ matrix of 8192-dimensional vectors paired with the binary correction outcome.

A leave-one-task-out logistic probe trained to predict correction from the embedding achieves $0.573$ accuracy, exactly matching the majority-class baseline. The linear probe is null: the final-layer mean-pooled embedding does not linearly carry the per-trial outcome. A weaker positive signal remains. The per-task cosine distance from \cond{L0\_self} is $\sim 0.02$ for all four relabel conditions, indicating that the relabel does shift the representation. The two external roles closest in representation are \cond{L\_memory} and \cond{L\_tool} (centroid distance $0.005$); the remaining external-role pairs separate by $0.009$ to $0.012$, comparable to the \cond{L0\_self}-to-external centroid distances.

The representation does shift under the relabel, but localizing that change would need a deeper protocol, per-layer or activation-patching with the model loaded in a framework with adequate GPU memory, which our setup lacked; we report this as a partial mechanistic finding.

\subsection{Within-Thought Duplication and Recency Control}
\label{app:mechanism-recency}

The \cond{L\_memory} condition appends a byte-identical copy of \cstar{} as a new \role{system} \role{<memory>} message, so the second copy of \cstar{} both raises its count in the prompt (duplication) and occupies the most recent message position before the audit (recency). Either confound, rather than the role tag, could in principle drive the $+53.3$\,pp main increase on Qwen-72B math. We isolate the role-tag claim with a three-way comparison that holds duplication count and recency position constant while varying only the role of the second copy.

\smallskip
\noindent\textbf{\cond{D0\_baseline}:} audit-only; \cstar{} appears once, inside the original \role{<thought>} block (equivalent to \cond{L0\_self}).

\noindent\textbf{\cond{D\_inthought\_dup}:} a byte-identical second copy of \cstar{} is appended to the assistant turn as \role{<thought>}\cstar\role{</thought>}, so \cstar{} now appears twice, both inside \role{<thought>}, with the second copy occupying the same temporal position as \cond{L\_memory}'s appended message.

\noindent\textbf{\cond{D\_memory\_ref}:} the canonical \cond{L\_memory} condition, re-run inside this batch for paired comparison; \cstar{} appears twice, once in \role{<thought>} and once in \role{system} \role{<memory>}.

\smallskip
\noindent The number of \cstar{} appearances ($1$, $2$, $2$), the position of the second appearance (none, end-of-trajectory, end-of-trajectory), and the byte content of every copy are matched between \cond{D\_inthought\_dup} and \cond{D\_memory\_ref}; the only variable is the role of the second copy.

\begin{table}[t]
\centering
\small
\setlength{\tabcolsep}{4pt}
\caption{Within-thought duplication and recency control on the Qwen-72B math failure pool ($n=30$ paired). $\Delta$ values are paired contrasts vs.\ \cond{D0\_baseline}; $p$ is from a two-sided $10{,}000$-sample paired bootstrap.}
\label{tab:appendix-recency-role}
\begin{tabular*}{\columnwidth}{@{\extracolsep{\fill}} p{0.56\columnwidth} r r @{}}
\toprule
\textbf{Condition} & CR & $\Delta$ vs.\ \cond{D0} \\
\midrule
\cond{D0\_baseline} (audit only)                              & 16.7\%          & --- \\
\cond{D\_inthought\_dup} (\cstar{} twice in \role{<thought>}) & 23.3\%          & $+6.7\;(p{=}0.26)$ \\
\cond{D\_memory\_ref} (canonical \cond{L\_memory})            & \textbf{70.0\%} & $\mathbf{+53.3^{***}}$ \\
\midrule
\multicolumn{3}{@{}p{\columnwidth}@{}}{\textit{Key contrast:} \cond{D\_memory\_ref} $-$ \cond{D\_inthought\_dup} $= +46.7$\,pp ($p{<}0.001$).} \\
\bottomrule
\end{tabular*}
\end{table}

\begin{figure*}[t]
\centering
\includegraphics[width=0.98\textwidth]{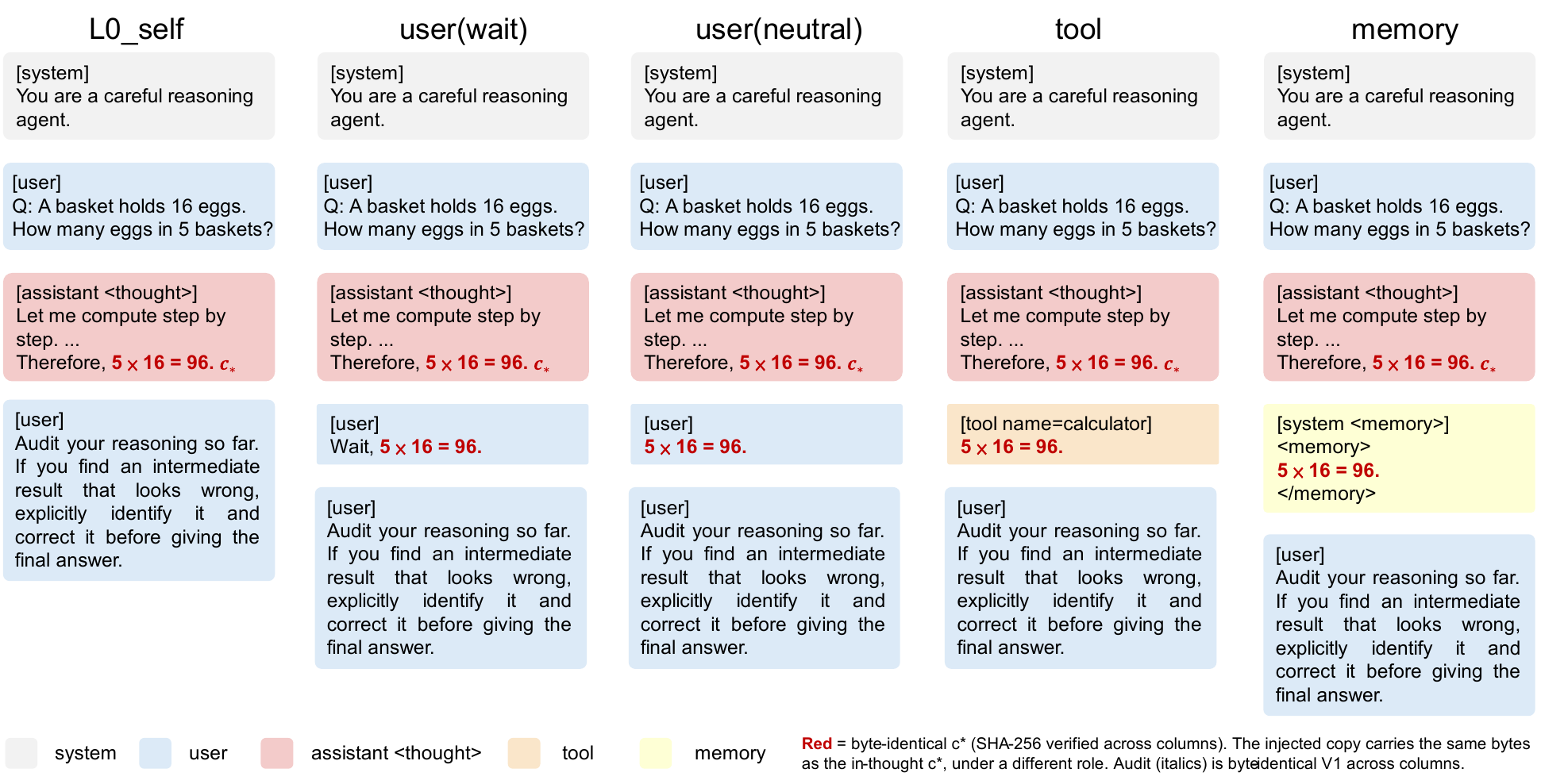}
\caption{The five chat-template conditions rendered verbatim. Red text is the byte-identical \cstar{} (SHA-256 verified); the audit (italics, bottom) is byte-identical V1 across columns.}
\label{fig:prompt-templates}
\end{figure*}

Three observations follow. First, duplicating \cstar{} inside \role{<thought>} alone produces only a $+6.7$\,pp shift over the audit-only baseline (CI $[0.0,\,+16.7]$, $p=0.26$), which is neither distinguishable from zero nor remotely comparable to the canonical $+53.3$\,pp increase. Second, the canonical \cond{L\_memory} value of $70.0\%$ reproduces the main-table number within this batch, providing an internal consistency check. Third, and most importantly, the paired contrast between \cond{D\_memory\_ref} and \cond{D\_inthought\_dup} is $+46.7$\,pp ($p<0.001$, CI $[+26.7,\,+66.7]$); since these two conditions are matched on duplication count, recency position, and byte content, this contrast isolates the pure role-tag effect. A duplication- or salience-only account of the main increase is incompatible with these numbers, and any residual recency contribution can supply at most $+6.7$\,pp.

%%%%%%%%%%%%%%%%%%%%%%%%%%%%%%%%%%%%%%%%%%%%%%%%%%%%%%%%%%%%%%%%%%%%%%%
%%%%%%%%%%%%%%%%%%%%%%% D: Reasoning-model scope %%%%%%%%%%%%%%%%%%%%%%
%%%%%%%%%%%%%%%%%%%%%%%%%%%%%%%%%%%%%%%%%%%%%%%%%%%%%%%%%%%%%%%%%%%%%%%
\section{Reasoning Models and Training-Based Self-Correction}
\label{app:reasoning}
Table~\ref{tab:final-main} identifies two experiments where the role-relabel does not add (gpt-oss-20B at $L_0=76.7\%$ and Qwen-72B BBH-LD at $L_0=66.7\%$); the former is reasoning-tuned, matching \citet{tsuiSelfCorrectionBench2025}, who observes a reduced or absent blind spot in reasoning-tuned and RL-trained models. Reasoning tuning alone is not sufficient: Qwen3-30B in reasoning mode retains $L_0=53.3\%$ and still gains $+30$\,pp under \cond{L\_memory}. We add a further data point with DeepSeek-R1 (served via GitHub Models), which emits explicit \role{<think>$\ldots$</think>} traces before its final answer. Free-tier rate limits cap per-condition $n$ to $10$--$13$ (57 trials across five conditions), but every trial is correct, a per-condition CR of $100\%$; since this is the strict-identification criterion used throughout, under audit-only DeepSeek-R1 explicitly names and rejects \cstar{} rather than re-deriving the answer silently. Where these RL-trained models close the final-answer gap by training (their stated objective), they close the strict-identification gap too, and the role-relabel reaches that same explicit outcome through prompt structure rather than training. DeepSeek-R1 reaches the ceiling under audit-only alone, as does GPT-4.1 in Appendix~\ref{app:baselines} ($L_0=82.4\%$). The scope is therefore models whose audit-only baseline is not saturated, chiefly standard instruction-tuned ones: a model already investing deliberate verification under audit-only addresses its own intermediate as a verifiable object, so the relabel adds nothing.

This delimits our relation to a parallel line that improves self-correction at training time rather than through the harness: multi-turn online RL on self-generated traces~\citep{kumarTrainingLanguageModels2024}, RL for self-verification and revision~\citep{maS2RTeaching2025}, and large-scale RL eliciting spontaneous self-checking~\citep{deepseekaiDeepSeekR12025}. First, the objectives differ. These methods optimize \emph{final-answer} self-correction, whether a wrong first attempt becomes a right one; ours is \emph{explicit identification}, whether the agent names and rejects the specific wrong intermediate. Appendix~\ref{app:final-answer-metric} shows the two come apart: the audit-only baseline already answers correctly in $70$ to $77\%$ of trials at a strict CR of $0$ to $17\%$, so a model whose final-answer correction is trained in can still lack addressability, the property a pipeline needs when an error must be surfaced and logged rather than silently overwritten. Second, the mechanism converts into a training signal rather than competing with one: since the lever is purely structural, a lightweight objective tagging thought-internal spans as self-addressable (Appendix~\ref{app:future}) could internalize the handle without an external role, identifying what to teach, namely addressability, not that more correction data helps. Third, the account also predicts the null results: where RL-trained reasoning models already invest verification under audit-only, they address their own intermediate as a verifiable object, so the relabel adds nothing. The contribution is thus a mechanistic explanation of why the gap exists in standard instruction-tuned models, and a zero-training diagnostic valid for the large deployed population of such models, rather than a competitor to training-time methods on reasoning models.

%%%%%%%%%%%%%%%%%%%%%%%%%%%%%%%%%%%%%%%%%%%%%%%%%%%%%%%%%%%%%%%%%%%%%%%
%%%%%%%%%%%%%%%% E: Chat-Template and Prompt Templates %%%%%%%%%%%%%%%%
%%%%%%%%%%%%%%%%%%%%%%%%%%%%%%%%%%%%%%%%%%%%%%%%%%%%%%%%%%%%%%%%%%%%%%%

\section{Chat-Template and Prompt Templates}
\label{app:chat-templates}

Role-token implementation differs substantially across model families: the same logical role (\role{system}, \role{user}, \role{tool}) is realized by entirely different special-token sequences set by each model's training-time chat template. This heterogeneity is precisely why the cross-family coverage of Table~\ref{tab:final-main} matters, since an increase surviving such varied tokenizations cannot be an artifact of any single template, and it is also why the role-handle abstraction must be documented at the token level rather than assumed uniform. We therefore record the exact begin/end role-tag pairs used for every model, so that the byte-identity guarantee of Sec.~\ref{sec:byte-identity} is auditable to the template boundary characters.

Figure~\ref{fig:prompt-templates} renders all five conditions verbatim on a representative GSM8K-style task with \cstar{} ``$5 \times 16 = 96$'', laying the columns side by side so that the byte-identical placement of \cstar{} across conditions is visually inspectable at a glance. The wrapping role is the only thing that changes from column to column; \cstar{} and the trailing audit stay constant, and the figure makes the single manipulated variable, the role tag, immediately legible against everything that is held fixed.

\paragraph{Per-model role tokens.} Each open-weight model is served via Ollama with its native chat template, so the role-tag tokenization matches the model's training-time format. Begin/end role-tag pairs are:

\smallskip
\noindent\small
\textbf{Qwen2.5-72B, Qwen3-30B:}\\
\texttt{<|im\_start|>\{role\}\textbackslash n} $\cdots$ \texttt{<|im\_end|>}\\[2pt]
\textbf{Phi-4-14B:}\\
\texttt{<|im\_start|>\{role\}<|im\_sep|>} $\cdots$ \texttt{<|im\_end|>}\\[2pt]
\textbf{Llama-3.3-70B:}\\
\texttt{<|start\_header\_id|>\{role\}<|end\_header\_id|>}\\
\hspace*{1em}$\cdots$ \texttt{<|eot\_id|>}\\[2pt]
\textbf{Gemma-3-12B:}\\
\texttt{<start\_of\_turn>\{role\}} $\cdots$ \texttt{<end\_of\_turn>}; its native template exposes only \texttt{user} and \texttt{model} roles, so a \role{system} message renders as a \texttt{user} turn and a \role{tool} message is dropped. On Gemma-3 this collapses \cond{L\_memory} onto \cond{L\_user\_neutral} (both $96.7\%$) and reduces \cond{L\_tool} to the audit-only baseline ($30.0\%$, with \cstar{} unrendered); its row therefore reflects this template limitation, not a null role effect.\\[2pt]
\textbf{gpt-oss-20B (harmony):}\\
\texttt{<|start|>\{role\}<|message|>} $\cdots$ \texttt{<|end|>}

\smallskip
\normalsize
\noindent Closed-weight models accept role strings via their JSON API surfaces (\texttt{role:} ``system''/``user''/``assistant''/``tool''); there the SHA-256 check is asserted on the \cstar{} payload string rather than the API-internal tokenization, which is not exposed, but the role-handle abstraction is preserved.

\paragraph{Tool-role realization.} The \cond{L\_tool} condition is rendered with a \texttt{name=``calculator''} field so that the tool-return is bound to a named tool rather than to an anonymous role. On Ollama-served families, this materializes as a \role{tool} message with its template-specific begin/end pair and a \texttt{name} field; on the closed-weight APIs, it materializes as the API's native tool-result message. The semantic intent is the same across templates: surface \cstar{} as if a calculator had just returned it.

\paragraph{Audit, byte-identity, and reproducibility.} The trailing audit is the V1 string of Table~\ref{tab:audit-paraphrase}; V2--V5 are exercised only in the audit-wording robustness check of \S\ref{sec:exp-robustness}. The implementation pipeline, including hashing, condition assembly, judge calls, and trial logging, follows the open-source code released with the paper; per-condition prompts can be reproduced bit-for-bit by running the released \texttt{exp\_relabel.py} (open-weight) and \texttt{exp\_relabel\_frontier.py} (closed-weight) drivers.

%%%%%%%%%%%%%%%%%%%%%%%%%%%%%%%%%%%%%%%%%%%%%%%%%%%%%%%%%%%%%%%%%%%%%%%
%%%%%%%%%%%%%%%%%%%%%%%%%%% F: Future Directions %%%%%%%%%%%%%%%%%%%%%%
%%%%%%%%%%%%%%%%%%%%%%%%%%%%%%%%%%%%%%%%%%%%%%%%%%%%%%%%%%%%%%%%%%%%%%%
\section{Future Directions}
\label{app:future}
The addressability account opens several directions. A per-task analysis correlating flips with properties of \cstar{}, its length, its distance from the audit, and the number of competing intermediates, would resolve the mechanism within each experiment, and a role-randomization control shuffling tags across the conversation would separate the injected handle from any global role-priority shift. Both extend existing machinery: the released logs for the former, the byte-identity assertion for the latter. On safety, the bare-injection adversarial mirror is the conservative end of a spectrum: pairing narrative-coherent payloads with calculator availability would identify the threshold where trust framing dominates, connecting our null result to the memory-poisoning literature. Three longer-horizon directions follow. First, the account predicts transfer to any domain whose intermediates can be named as discrete objects, so extending the relabel to structured artifacts, a buggy diff or faulty planning step, tests it at scale. Second, self-addressability may substitute for an external role: an inline within-\role{<thought>} span tag named in the audit tests this at the prompt level, and a lightweight training signal tagging such spans could internalize the handle; the duplication control ($+6.7$\,pp, Appendix~\ref{app:mechanism-recency}) suggests pointing without a role change falls short. Because the lever is structural, such a signal would convert a harness-level intervention into a training-level property. Third, the explicit-correction gain at no significant final-answer cost (Appendix~\ref{app:final-answer-metric}) defines a design space for harnesses: canonical handle phrases and a verify-before-accept default on \role{user} and \role{system <memory>} content are starting policies. We release a benchmark split with hashed \cstar{} placements, alongside the code in Appendix~\ref{app:chat-templates}, so the effect can be remeasured on future models under byte-identical conditions.

\section{The Localization Premise}
\label{app:localization}
The protocol injects a known \cstar{} and measures whether relabeling it elicits explicit correction; by the failure-pool construction (Sec.~\ref{sec:eval-stats}), \cstar{} is given. This isolates the \emph{relabel} step but brackets a harder one, \emph{localizing} which intermediate is wrong, the binding constraint: models repair an error once its location is supplied but cannot reliably find it themselves~\citep{tyenLLMsCannotFind2024}. We therefore frame deployment as a two-stage pipeline: an upstream detector proposes candidate spans, and the relabel lever acts on each. The present paper establishes only the second stage, and any end-to-end ``production'' claim would overstate the result. Two observations bound how demanding the detector must be. First, candidates can come from cheap, training-free sources already common in harnesses, self-consistency disagreement, a separate verifier or tool re-check, or sanity-check failures; the lever needs candidates, not a perfect localizer. Second, the adversarial mirror (Sec.~\ref{sec:exp-adversarial}) bounds the cost of false positives: externally presented claims are adopted at most $3.3\%$ of the time, so relabeling a \emph{correct} span is unlikely to induce a spurious correction, though we did not test this directly since the failure pool holds only wrong intermediates. Recording that false-correction rate is a direct extension of the byte-identity machinery and the natural next experiment.
\end{document}